\pdfoutput=1
\documentclass[11pt]{article}
\usepackage{multirow}

\usepackage[preprint]{acl}

\usepackage{times}
\usepackage{latexsym}
\usepackage{amsmath}
\usepackage{tcolorbox}
\usepackage{booktabs}
\usepackage[nameinlink]{cleveref}
\usepackage{subcaption}
\usepackage{placeins}

\usepackage[T1]{fontenc}

\usepackage[utf8]{inputenc}

\usepackage{microtype}

\usepackage{inconsolata}

\usepackage{graphicx}

\usepackage{xspace}
\usepackage{colortbl}
\usepackage{xcolor}
\usepackage{makecell}

\definecolor{relColor}{RGB}{255, 230, 230}   
\definecolor{genColor}{RGB}{255, 245, 220}  
\definecolor{locColor}{RGB}{230, 255, 230}   
\definecolor{portColor}{RGB}{230, 245, 255}   

\newcommand{\dataset}{CulturalBind\xspace}

\title{Evaluating Cultural Knowledge Editing in Vision-Language Models}

\author{
    {\bf Malvina Nikandrou}$^{\dagger}$
    {\bf \enspace Georgios Pantazopoulos}$^{\dagger}$
    {\bf \enspace Ioannis Konstas}$^{*}$
    {\bf \enspace Alessandro Suglia}$^{\dagger}$\\
    \AND \textnormal{$^\dagger$University of Edinburgh\ \ $^*$Heriot-Watt University}\\
}

\usepackage{enumitem}

\begin{document}
\maketitle
\begin{abstract}
Current Vision-Language Models (VLMs) frequently fail to recognize culture-specific concepts, especially from underrepresented regions.
Effectively injecting this long-tail knowledge into existing models remains a challenge. In this work, we investigate knowledge editing as a targeted alternative by introducing CulturalBind, a benchmark comprising 3.2K images of 1K concepts from 6 domains, designed to evaluate culture-aware cross-modal knowledge as well as concept-level knowledge editing in VLMs.
Pre-edit evaluation reveals that textual and visual knowledge is decoupled: models often demonstrate text-based knowledge of cultural concepts but fail to identify them from images. We then evaluate editing methods in single and sequential scenarios. Although current methods successfully correct individual visual recognition errors, they exhibit low cross-modal portability, failing to integrate edits into the model's broader parametric knowledge. Furthermore, performance degrades rapidly under sequential editing, highlighting the need for comprehensive adaptation approaches that better consolidate multimodal knowledge over time.
\end{abstract}

\section{Introduction}

Current Vision-Language models (VLMs) exhibit broad visual and linguistic capabilities \citep{bai2025qwen3vltechnicalreport, gemmateam2025gemma3technicalreport, Wang2025InternVL35AO}, yet frequently fail to recognize culture-specific concepts such as traditional musical instruments, regional dishes, and local festivals, particularly those from underrepresented cultures. This reflects a systematic imbalance in pretraining corpora that fail to capture the long tail of global cultural knowledge \citep{vayani2024alm, richards2024does, nyandwi-etal-2025-grounding}. As VLMs are increasingly deployed in global contexts, this limitation risks perpetuating cultural blind spots.

One approach to addressing this cultural gap is by curating more diverse pretraining datasets~\cite{penedo2024fineweb, Albalak2024:llm_data}.
However, improvements in long-tail cultural concepts are expected to diminish at scale \citep{hooker2025slow, NEURIPS2024_715b78cc}, making full retraining an expensive and uncertain fix.
Moreover, subpar performance on cultural benchmarks may not reflect only limited data coverage.
Work on general entity knowledge has shown that VLMs can showcase factual knowledge, and yet fail to access it through visual input \citep{ashok2025can, cohen-etal-2025-performance}, suggesting a similar modality gap may underlie cultural recognition failures.

Knowledge editing offers an efficient alternative, where localized parameter updates address gaps identified post training without affecting the model's existing knowledge \citep{wang2024knowledge}.
Unlike prior work on entity-level editing \citep{huang2024vlkeb, li-etal-2024-mike,zhang-etal-2025-mc}, where the target is a unique referent, we target concept-level knowledge: the goal is to acquire a generalizable visual category that transfers across instances of the same concept and integrates with the model's broader parametric knowledge.
Whereas an entity edit is validated on the same referent under varying conditions (e.g., the Golden Gate Bridge from different viewpoints), a concept edit must transfer to different instances of the category (e.g., nadaswarams of different makes, photographed in different settings).

\begin{figure*}[h!]
   \centering
    \includegraphics[width=\textwidth]{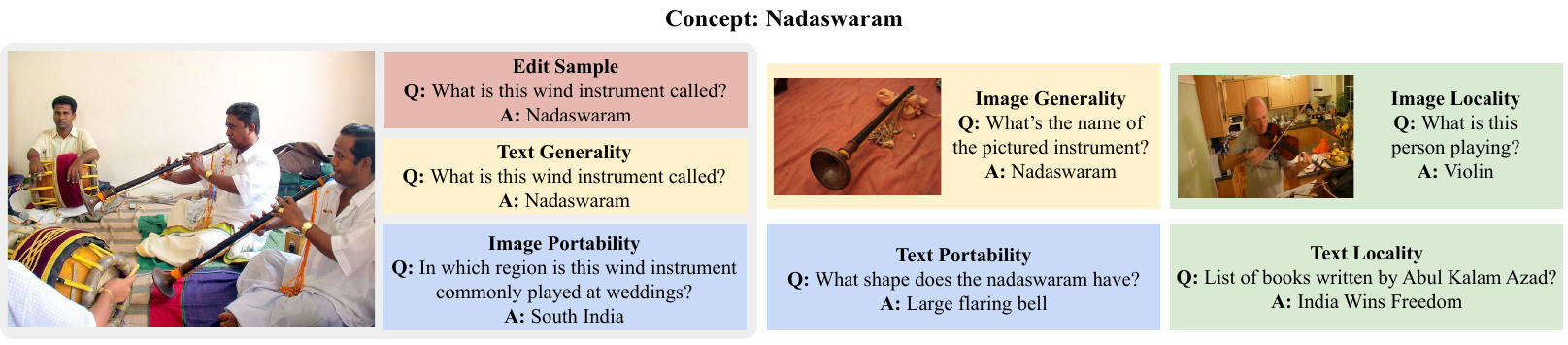}
    \caption{Structure of the \dataset editing dataset. Each concept is paired with an edit sample and six auxiliary probes targeting generalization (Text and Image Generality), knowledge preservation (Text and Image Locality), and cross-modal knowledge integration (Text and Image Portability).}
    \label{fig:dataset}
\end{figure*}

To study this problem, we construct \dataset, a cross-modal evaluation benchmark of 1K concepts spanning 14 regions and 6 cultural domains, which pairs each concept with encyclopedic and visual attribute knowledge documented in Wikipedia.
Before editing, we find that models demonstrate moderate parametric knowledge when probed through language, but recognition from images lags by large margins that do not close with scale.
The correlation between textual recall and visual recognition is weak, indicating that the two are largely decoupled.
This indicates that the bottleneck is not the absence of factual knowledge but the failure to access it through visual referents.

To evaluate whether editing methods can support cultural adaptation, we assess a range of parameter-modifying and memory-based approaches under two editing scenarios. Single edits evaluate whether methods can fix individual recognition failures, while sequential edits reflect the more practical setting where a model must acquire new concepts incrementally over time. Both settings are evaluated across standard editing metrics and two portability dimensions specific to cross-modal knowledge integration (Figure~\ref{fig:dataset}):
i) \textit{Text-Portability}, whether a recognition edit transfers to text-only recall of visual attributes, and ii) \textit{Image-Portability}, whether it enables visually grounded access to encyclopedic knowledge.

Results demonstrate that while editing methods reliably achieve recognition, portability remains uniformly low: editing establishes associations between images and concept labels without grounding them in the model's broader parametric knowledge, indicating that the visual-linguistic binding gap identified in the cross-modal evaluation persists after editing. In the sequential editing setting, this limitation is exacerbated by rapid performance degradation.
Overall, these results indicate that cultural concept acquisition remains a challenge that requires more comprehensive continual adaptation techniques that consolidate multimodal knowledge.

\section{Related Work}

\paragraph{Cultural Evaluation of VLMs}
Recent work has expanded the scope of cultural evaluation for VLMs, including largely multilingual benchmarks \citep{vayani2024alm, romero2024cvqa} and diverse task settings \citep{thapliyal-etal-2022-crossmodal3600, yin2021gd-vcr, richards2024does, bhatia-etal-2024-local}.
Benchmarks probing cultural knowledge typically use VQA, where a correct answer requires both visual recognition and factual recall.
CVQA \citep{romero2024cvqa} and CulturalVQA \citep{nayak2024benchmarking} do not pair these questions with text-only counterparts, so the two failure sources cannot be separated, while MMA-ASIA \citep{zheng2025mmaasia} pairs text-only and visual versions of the same questions.
\dataset extends this pairing across distinct knowledge dimensions.

\paragraph{Cultural Adaptation of VLMs}
Fewer works have explored approaches to improving cultural awareness in VLMs at different stages of model development.
Prior work has proposed geo-diverse pretext objectives \citep{46553}, data augmentation through code-switching and image editing \citep{li-zhang-2023-cultural}, and interventions in data compositions \citep{ignat-etal-2024-annotations-budget} during pretraining.
At the post-training level, recent work has explored pipelines for culturally-aware data construction \citep{nyandwi-etal-2025-grounding}.
At inference time, retrieval-augmented approaches have been proposed to inject cultural knowledge without modifying model parameters \citep{anugraha2025m4, ibrahim2025concap, nikandrou2024crope, park-etal-2025-evaluating}.
These methods target different points in the model lifecycle and can be considered as complementary strategies to improve cultural awareness of VLMs.
In this work, we explore knowledge editing as a potential direction for injecting cultural knowledge through parameter updates that are computationally efficient and localized by design.

\begin{table*}[t]
\centering
\footnotesize \setlength{\tabcolsep}{4pt}
\begin{tabular}{llll}
\toprule
\textbf{Dimension} & \textbf{Visual Input} & \textbf{Text Input} & \textbf{Tests} \\
\midrule
Reliability         & Edit image   & Recognition question          & Edit success \\
Text-Generality     & Edit image   & Paraphrased question          & Text generalization \\
Image-Generality    & Held-out image & Recognition question        & Image generalization \\
\midrule
Text-Locality       & —            & Unrelated text question          & Preservation of text knowledge \\
Image-Locality      & Unrelated image & Unrelated visual question     & Preservation of VL knowledge \\
\midrule
Text-Portability    & —            & Visual attribute question  &  Textual attribute recall  \\
Image-Portability   & Edit image   & Encyclopedic question  & Visually-grounded factual recall \\
\bottomrule
\end{tabular}
\caption{Evaluation dimensions for knowledge editing. Generality targets the generalization of the edit, Locality focuses preservation of unrelated knowledge, and Portability tests cross-modal knowledge integration.}
\label{tab:eval_dimensions}
\end{table*}

\paragraph{Multimodal Editing Evaluations}
Early work by \citet{cheng2023can} expanded text knowledge editing frameworks to captioning and VQA tasks, but relied on synthetic images for evaluating edit generalization and offered no assessment of portability.
Subsequent work has extended this framework in several directions.
VLKEB \citep{huang2024vlkeb} extends the standard reliability, locality, and generality metrics with a portability dimension. MIKE \citep{li-etal-2024-mike} focuses on fine-grained entity recognition and introduces a multi-task editing evaluation. MC-MKE \citep{zhang-etal-2025-mc} defines multimodal knowledge in terms of visual and textual triples and evaluates portability from both text and visual edits.
More recently, MMKE-Bench \citep{dummke} broadens the scope beyond structured knowledge triplets and includes visual semantic and user-specific editing scenarios.
Rather than counterfactual edits that overwrite accurate associations, we focus on concept acquisition: the model learns visual categories that it previously failed to recognize.  
Edit images in \dataset are also less visually similar to their generality images than those in entity-based benchmarks (\Cref{app:entity_vs_concept}), reflecting the greater visual variation across instances of a concept.
Additionally, while existing benchmarks assess whether encyclopedic knowledge is accessible from visual input after editing \citep{huang2024vlkeb}, we further evaluate whether the model acquires descriptive knowledge of a concept's visual attributes accessible through text alone.

\section{Knowledge Editing Formulation}
Knowledge editing modifies a specific  piece of knowledge $k$ within a model with parameters $\theta_{t-1}$ through an editing process $E_{\phi}$:

\begin{equation} \label{eq1}
\begin{gathered}
\theta_t = E_{\phi}(\theta_{t-1}, k) \\
\theta_t(k') = \theta_{t-1}(k') \quad \forall k' \neq k
\end{gathered}
\end{equation}

Each edit sample $k = (v, q, a^*, \hat{a})$ consists of an input image $v$, a recognition question $q$ (e.g., \textit{``What is this instrument called?''}), the correct answer $a^*$, and the model's incorrect pre-edit answer $\hat{a}$.
Unlike factual editing, where the target is a specific entity or proposition, we target \emph{concept-level} knowledge \citep{wang2024editing}: rather than memorizing a single visual exemplar, the goal is to establish a concept-level association that generalizes across instances of the same category.

\Cref{tab:eval_dimensions} summarizes the evaluation dimensions.
\textbf{Reliability} measures edit success, while \textbf{Text-Generality} and \textbf{Image-Generality} capture generalization to paraphrases of the input question or held-out images of the same concept, respectively.
 \textbf{Text-Locality} and \textbf{Image-Locality} measure preservation of unrelated text-only and multimodal knowledge.
Portability evaluates whether the edited model can apply the acquired knowledge beyond recognition. We focus on cross-modal portability:
\textbf{Text-Portability} probes for knowledge transfer by testing whether a model that has learned to visually recognize a concept can also recall its visual attributes without image input.
\textbf{Image-Portability} assesses whether recognition can be combined with encyclopedic factual recall through visually grounded questions, requiring the model to bind the concept from an image with factual knowledge.
\Cref{fig:dataset} demonstrates an example across evaluation dimensions.

\section{\dataset Edit Dataset}\label{sec:ke_data}

We construct a dataset to test VLMs' knowledge of cultural concepts that refer to generalizable categories with multiple real-world instances, as opposed to named entities, whose recognition requires memorizing a unique referent.
The full resource, \textit{\dataset}, comprises 3.2K images from 1K concepts for cross-modal knowledge evaluation (\Cref{sec:cross_modal}).
For editing experiments, we derive the editing samples by filtering \dataset based on model failures, and pairing samples with text and image locality probes (\Cref{sec:ke_experiments}).

\subsection{\dataset Construction}

\paragraph{Concept Selection}
We collect concepts from two sources: 1) CulturalGround \citep{nyandwi-etal-2025-grounding} is a large-scale cultural VQA dataset covering 42 countries and sourced from Wikidata, and 2)  CROPE \citep{nikandrou2024crope}, a smaller probing dataset containing culture-specific concepts across five regions validated by native-speaker annotators.

\paragraph{Image Collection and Filtering}
Evaluating generality across images requires at least two images per concept.
For concepts with a single image, we retrieve up to 10 Creative Commons-licensed images using the Openverse API\footnote{\url{https://api.openverse.org/v1/}}.
Perceptual hashing\footnote{Using the Python library \href{https://github.com/JohannesBuchner/imagehash}{imagehash}.} is applied to deduplicate across all images.
We then apply a three-stage filtering process to ensure that retained images are relevant and unambiguous.
First, we use GPT-5.2\footnote{gpt-5.2-2025-12-11.} to dismiss images that contain the concept name in text, thereby preventing VLMs from bypassing visual recognition.
Second, GPT-5.2 scores each image against the concept's Wikipedia summary to retain only representative images.
Third, we manually inspect all retained Openverse and CulturalGround images to remove remaining ambiguous cases, discarding 96 images.
Concepts with fewer than two images are excluded.

\paragraph{Knowledge QA Generation}
We extract two types of questions from each concept's Wikipedia page:
\textit{visual attribute} questions, covering features observable in images, and \textit{encyclopedic} questions, covering knowledge that cannot be inferred from images alone.
\Cref{tab:background_questions} reports the distribution across question categories.
For visual attribute knowledge, we prompt GPT-5.2 to extract visual descriptions explicitly stated in the Wikipedia text across three observable categories:
physical attributes, actions, and contextual cues, yielding 5,321 questions converted into QA format.

For encyclopedic knowledge, we prompt GPT-5.2 to generate question-answer pairs from a Wikipedia text using a taxonomy of eight knowledge categories (\Cref{tab:background_questions}), excluding information present in the visual descriptions.
To evaluate \textit{visually-grounded encyclopedic QA}, the concept name in each question is replaced with a referring expression, requiring the model to identify the concept from an image before retrieving the associated background knowledge.
Following~\Cref{fig:dataset}, ``In which region is the \underline{nadaswaram} commonly played at weddings?'' becomes ``In which region is \underline{this wind instrument} commonly played at weddings?''. We filter out questions that GPT-5.2 can answer correctly without images, resulting in a total of 5,298 encyclopedic questions.
We manually audit a sample of 200 QA pairs for question clarity and answer faithfulness (\Cref{app:qa_audit}).

\paragraph{Dataset Summary}

\begin{table}
\centering
\begin{minipage}[t]{0.5\textwidth}
\centering
\small\setlength{\tabcolsep}{2pt}
\begin{tabular}[t]{lrr}
\toprule
Region & Concepts & Samples \\
\midrule
Eastern Asia             & 199 & 571 \\
South-eastern Asia       & 164 & 533 \\
Southern Asia            & 168 & 519 \\
Western \& Central Asia  & 118 & 366 \\
Southern Europe          & 149 & 356 \\
Northern Europe          & 81 & 166 \\
Eastern Africa           &  22 & 136 \\
Eastern Europe           & 45 & 108 \\
Western Europe           &  26 & 100 \\
Central Am.\ \& Caribbean &  33 & 92 \\
Northern America         &  19 &  85 \\
Western Africa           &  24 &  64 \\
Southern America         &  23 & 62 \\
Northern Africa          & 19 & 58 \\
\midrule
Total                    & 1{,}090 & 3{,}216 \\
\bottomrule
\end{tabular}
\caption{\dataset distribution by region.}\label{tab:dataset_region}
\end{minipage}
\end{table}
The final dataset comprises 1,090 unique concepts paired with 3,216 images, spanning 14 geographic regions and 6 cultural domains.
The number of concepts and total samples per region are listed in \Cref{tab:dataset_region}.
Geographically, coverage is broadest in Asian regions, reflecting the cultural diversity of the source datasets, shown in \Cref{fig:geo_map}.
African sub-regions are underrepresented, mirroring known gaps in Wikidata and Wikipedia coverage.
Samples across America are collectively sparse for different reasons: North American concepts are de-prioritized in long-tail cultural evaluations, while Central and South American regions reflect sparser Wikidata coverage.
By domain (see \Cref{tab:dataset_domain}), Foods \& Beverages account for nearly 40\% of concepts, while the remaining domains are broadly balanced across concrete categories and human activities.

\begin{table*}[t]
\centering
\setlength{\tabcolsep}{5pt}
\small
\begin{tabular}{lccc|cc}
\toprule
 & \multicolumn{3}{c}{\textbf{Text-Based Metrics}} & \multicolumn{2}{c}{\textbf{Image-Based Metrics}} \\\midrule
Model & OE Description & Encyclopedic & Attribute & Recognition & VG Encyclopedic \\\midrule
\addlinespace
InternVL3.5 (4B)  & 35.7 & 24.8 & 31.6 & 18.6 & 15.5 \\
InternVL3.5 (8B)  & 46.2 & 30.0 & 37.2 & 18.7 & 16.4 \\
InternVL3.5 (14B) & 54.0 & 33.1 & 42.3 & 19.8 & 17.3 \\
InternVL3.5 (38B) & 60.9 & 38.9 & 47.3 & 25.7 & 20.1 \\
\addlinespace
Qwen3-VL (4B)  & 56.2 & 32.4 & 39.5 & 34.6 & 23.9 \\
Qwen3-VL (8B)  & 67.7 & 39.3 & 46.7 & 34.9 & 24.3 \\
Qwen3-VL (32B) & 77.4 & 47.1 & 55.7 & 36.3 & 28.1 \\
\addlinespace
Gemma 3 (4B)  & 54.2 & 31.1 & 37.3 & 31.9 & 21.2 \\
Gemma 3 (12B) & 73.8 & 43.1 & 50.6 & 38.2 & 26.8 \\
Gemma 3 (27B) & 82.8 & 52.2 & 56.8 & 43.1 & 33.7 \\
\addlinespace
GPT 5.2 & 90.5 & 67.7 & 69.3 & 44.6 & 40.2 \\
\bottomrule
\end{tabular}
\caption{Cross-modal knowledge evaluation. Text-based measures (Open-Ended Description, Encyclopedic QA, Attribute QA) probe parametric cultural knowledge via text, while image-based measures (Recognition,  Visually-Grounded Encyclopedic QA) test whether knowledge about the same concepts is accessible through visual input.}
\label{tab:preliminary}
\end{table*}
\subsection{Converting to KE Format}
Existing multimodal KE benchmarks construct edit samples either from model failures \citep{cheng2023can} or counterfactually \citep{huang2024vlkeb,dummke}, replacing the true label with a plausible but incorrect target.
Counterfactual construction produces a model-agnostic dataset suited to simulating knowledge updates, but can introduce logical inconsistencies: after relabeling images of a nadaswaram as ``oboe'', the expected answer for genuine oboe images becomes undefined.
By selecting edit samples only for concepts the model fails to recognize, the edit target is always factually correct and consistent with image content, grounding evaluation in practical knowledge acquisition scenarios.

Each edit sample is linked to auxiliary samples that evaluate metrics beyond reliability. For text generality, the original question is paraphrased using Qwen3-32B \citep{yang2025qwen3} and paired with the original image, while for image generality, a held-out image of the same concept is paired with the original question.
Following MMEdit \citep{cheng2023can}, text locality samples are drawn from Natural Questions \citep{kwiatkowski-etal-2019-natural}.
For image locality, VQA-v2 \citep{vqav2} samples are selected by averaging the image-image and text-text similarities with the edit sample using SigLIP2 embeddings \citep{tschannen2025siglip}.
For text portability, we use the generated visual attribute questions posed without image input, while for image portability, visually-grounded encyclopedic questions are paired with the edit image.

\section{Cross-Modal Knowledge Evaluation}\label{sec:cross_modal}
A concern when sourcing concepts from Wikipedia is that VLMs may already have parametric knowledge of them, given Wikipedia's prominence as a pretraining source \citep{penedo2024fineweb}.
However, even if a model has knowledge of a concept, it might fail to access it through visual inputs.
For example, a VLM may learn from text data that the nadaswaram is a wind instrument with a wooden body and a flaring bell, yet fail to identify one from an image.
Following recent work surfacing gaps between text-based and image-based performance on factual recall tasks \citep{ashok2025can, cohen-etal-2025-performance}, we investigate whether a similar disconnect underlies failures on cultural concepts.

To establish if a model has a baseline parametric knowledge of a concept, we evaluate \textit{Open-Ended Description}: given only a name, the model is asked to provide a short definition.
Text-based knowledge is further probed through the \textit{Encyclopedic QA} and \textit{Attribute QA} pairs in \dataset.
On the image-based side, \textit{Recognition QA} tests whether VLMs can identify the concepts from images, and \textit{Visually-Grounded Encyclopedic QA} additionally requires composing recognition with factual recall.

\subsection{Experimental Setup}

We evaluate open-weights models from three families: InternVL3.5 \citep{Wang2025InternVL35AO}, Qwen3-VL \citep{bai2025qwen3vltechnicalreport}, and Gemma 3 \citep{gemmateam2025gemma3technicalreport} with sizes between 4B and 38B parameters.
Additionally, we report the performance of GPT-5.2 as an upper-bound closed-source model.
All models are evaluated using greedy decoding to ensure deterministic, reproducible outputs.
Responses are evaluated using Qwen3-32B \citep{yang2025qwen3} as the LLM judge against reference answers ensuring future reproducibility.
Open-Ended Description responses are scored with a binary rubric, and all QA tasks are evaluated using a graded rubric adapted from \citet{manas2024lave}.
Following best practices~\cite{bavaresco2025llms}, we validate judge reliability against human scores, finding strong agreement  (average QWK = 0.86).
Further details provided in the \Cref{appendix:llm-judge}.

To assess the coupling between textual knowledge and visual access, we divide concepts into \textit{Known} and \textit{Unknown} groups based on their Open-Ended Description scores, and compare Visual Recognition accuracy across groups.
We measure the correlation between the binary Description and graded Visual Recognition scores using point-biserial correlation ($r_{pb}$).

\subsection{Results}

Table \ref{tab:preliminary} presents text-based and image-based performance across evaluated models.
Models achieve the highest performance on Open-Ended Description, with several models reaching scores above 70\%.
This suggests that models possess substantive basic knowledge of the examined cultural concepts.
Performance on Encyclopedic and Attribute QA drops considerably, indicating that background knowledge coverage is more limited.
Models score approximately 7 points higher on Attribute QA than Encyclopedic QA across smaller models, likely because salient perceptual properties such as shape, color, and material are more broadly represented in pretraining text than specific factual knowledge.
This advantage narrows for stronger models, as Encyclopedic QA improves.

All models struggle to identify the same concepts from images, with recognition scores consistently lagging behind text-based performance by a large margin.
Within model families, scaling improves text-based performance more than visual recognition: Gemma 3 4B$\rightarrow$27B gains approximately 29 points on Description but only 11 on Recognition, while for Qwen3-VL 4B$\rightarrow$32B, the corresponding gains are 21 and fewer than 2 points.
This suggests the modality gap is not merely a capacity limitation that larger models overcome.

Across models, the drop from text-only Encyclopedic QA to Visually-Grounded Encyclopedic QA ranges from approximately 8 to 28 points, with the largest gap observed for GPT-5.2 despite its overall strength.
This indicates that visual grounding is a bottleneck:
models may possess the relevant background knowledge but cannot access it from visual referents.
This aligns with \citet{cohen-etal-2025-performance}'s findings on knowledge extraction gaps in VLMs.
Results vary substantially across regions, while domain-level variance remains limited across all dimensions, as shown in \Cref{app:cross_eval}.

\begin{table}
    \centering
    \small
    \begin{tabular}{lcccl@{}}
    \toprule
    Model & Known & Unknown & $\Delta$ & $r_{pb}^*$ \\
    \midrule
    \addlinespace
    InternVL3.5 (4B)  & 25.8 &          10.8 &       15.0 & 0.267 \\
    InternVL3.5 (8B)  & 22.9 &          10.5 &       12.4 & 0.221 \\
    InternVL3.5 (14B) & 22.3  &          10.6 &       11.7 & 0.215 \\
    InternVL3.5 (38B) & 29.5  &          11.9 &       17.6 & 0.265 \\
    \addlinespace
    Qwen3-VL (4B)     & 42.4 &          15.9 &       26.5 & 0.374 \\
    Qwen3-VL (8B)     & 37.2 &          16.2 &       21.1 & 0.286 \\
    Qwen3-VL (32B)    & 35.5 &          15.2 &       20.3 & 0.241 \\
    \addlinespace
    Gemma 3 (4B)    & 37.9  &          16.6 &       21.3 & 0.314 \\
    Gemma 3 (12B)     & 37.9 &          17.2 &       20.7 & 0.256 \\
    Gemma 3 (27B)     & 39.8 &          24.0 &       15.8 & 0.158 \\
    \addlinespace
    GPT-5.2         & 41.0 &          18.8  &       22.2 & 0.177 \\

     \bottomrule
    \end{tabular}
 \caption{Recognition accuracy for correctly versus incorrectly
recalled concepts. \textit{Known}: model accurately described the
concept; \textit{Unknown}: model failed to describe it. Correlation ($r_{pb}$) indicates relationship strength. $^*p < .01$.}
\label{tab:analysis_corr}
\end{table}

Table~\ref{tab:analysis_corr} breaks down the Recognition accuracy grouped according to their Description scores.
Models consistently achieve higher recognition accuracy for known concepts with differences of up to 26.5 percentage points, and the correlation of Description and Recognition scores is statistically significant across all models.
However, correlations remain weak ($r_{pb}$ = .16–.37), indicating that textual recall and visual recognition are largely decoupled.
A model’s ability to generate a textual description for a cultural concept does not serve as a reliable predictor of its ability to identify it in an image.
We note again that the correlation does not strengthen with scale. Gemma 3-27B ($r_{pb}$ = .158) and GPT-5.2 ($r_{pb}$ = .177) show the weakest associations despite being among the strongest models overall.
These results suggest that current multimodal pretraining does not encourage cross-modal consistency, even as general text and visual capabilities improve.

\section{Knowledge Editing Evaluation}\label{sec:ke_experiments}

\begin{table*}
\centering
\small
    \begin{tabular}{ll
      >{\columncolor{relColor}}c   
  >{\columncolor{genColor}}c   
  >{\columncolor{genColor}}c   
  >{\columncolor{locColor}}c   
  >{\columncolor{locColor}}c   
  >{\columncolor{portColor}}c  
  >{\columncolor{portColor}}c  
    c}\toprule
    Model & Method & Reliability & T-Gen & I-Gen & T-Loc & I-Loc & T-Port & I-Port & Avg \\\midrule
 & FT-V & 98.5 & 79.7 & 25.1 & 93.0 & 77.7 & 15.7 & 16.0 & 58.0 \\
 & FT-T & 100.0 & 79.8 & 63.4 & 75.7 & 84.2 & 14.0 & 14.1 & 61.6 \\
 & DPO & 99.8 & 78.9 & 62.8 & 75.4 & 84.3 & 14.0 & 14.2 & 61.3 \\
Qwen3-VL (4B)  & IKE & 99.7 & 98.9 & 99.8 & 23.4 & 31.7 & 15.8 & 17.4 & 55.2 \\
N=871 & MEND & 92.3 & 88.1 & 85.9 & 86.1 & 82.3 & 12.7 & 12.5 & \textbf{65.7} \\
 & GRACE & 99.5 & 2.4 & 4.6 & 93.0 & 97.4 & 15.7 & 15.0 & 46.8 \\
& WISE-T & 100.0 & 97.7 & 98.1 & 93.0 & 38.4 & 7.2 & 8.0 & \underline{63.2} \\
& WISE-V & 83.8 & 85.5 & 79.6 & 92.9 & 72.9 & 12.7 & 10.5 & 62.6 \\
\midrule
 & FT-V & 98.7 & 79.8 & 22.6 & 93.6 & 78.2 & 21.3 & 20.6 & 59.3 \\
 & FT-T & 100.0 & 84.5 & 70.3 & 79.8 & 84.4 & 19.2 & 15.6 & 64.8 \\
 & DPO & 100.0 & 84.7 & 70.1 & 79.1 & 83.9 & 19.1 & 15.4 & 64.6 \\
Qwen3-VL (8B)  & IKE & 96.8 & 96.4 & 96.4 & 31.6 & 29.9 & 20.7 & 23.2 & 56.4 \\
N=864 & MEND & 93.1 & 89.2 & 85.9 & 89.6 & 80.8 & 17.8 & 13.8 & \underline{67.2} \\
 & GRACE & 100.0 & 2.0 & 6.4 & 93.6 & 99.2 & 21.3 & 17.0 & 48.5 \\
& WISE-T & 100.0 & 97.0 & 96.1 & 93.5 & 57.4 & 17.3 & 14.2 & \textbf{67.9} \\
& WISE-V & 86.6 & 86.5 & 78.2 & 93.6 & 79.8 & 19.3 & 15.2 & 65.6 \\
\midrule
 & FT-V & 99.4 & 87.8 & 22.5 & 96.7 & 73.2 & 22.0 & 20.7 & 60.3 \\
 & FT-T & 100.0 & 94.1 & 62.4 & 91.3 & 82.7 & 20.7 & 17.4 & 66.9 \\
 & DPO & 100.0 & 94.5 & 63.4 & 91.6 & 82.6 & 20.6 & 17.5 & \textbf{67.2} \\
Gemma 3 (12B)  & IKE & 99.4 & 99.1 & 99.4 & 30.5 & 26.5 & 21.2 & 28.1 & 57.8 \\
N=796 & MEND & 86.6 & 82.0 & 62.7 & 93.2 & 80.2 & 20.6 & 15.9 & \underline{63.0} \\
 & GRACE & 98.7 & 2.9 & 9.9 & 96.7 & 91.2 & 22.0 & 19.4 & 48.7 \\
& WISE-T & 99.5 & 97.6 & 88.7 & 70.1 & 44.0 & 12.1 & 7.6 & 59.9 \\
& WISE-V & 99.4 & 97.9 & 89.7 & 80.4 & 51.1 & 13.7 & 8.2 & 62.9 \\
\bottomrule
\end{tabular}
\caption{Method performance on single edits to correct N model errors on cultural concept recognition.}\label{tab:main_qwen_4}
\end{table*}

\subsection{Experimental Setup}

To address the identified recognition gap, we apply representative knowledge editing methods alongside finetuning baselines. We select Qwen3-VL (4B, 8B) and Gemma 3 (12B) as base models, given their stronger cross-modal performance.
As baselines, we apply finetuning of the module which connects the visual encoder to the language model (FT-V), and the final language model layer (FT-T) using standard cross-entropy on response tokens, and Direct Preference Optimization (DPO) \citep{rafailov2023direct} using the model's incorrect pre-edit predictions as negative samples.

We evaluate four knowledge editing methods: IKE \citep{zheng2023can}, an in-context editing approach that retrieves
demonstration examples at inference time without modifying model parameters;
MEND \citep{mitchell2021fast}, a meta-learning approach that trains a hypernetwork to predict parameter updates for new edits; GRACE \citep{hartvigsen2023aging}, a memory-based approach that stores edited representations in a learned codebook; and WISE \citep{wang2024wise}, a dual-memory approach that routes inputs between original and edited parameter
sets. For WISE, we implement two routing variants:
text-based locality following the original formulation (WISE-T), and image-based locality (WISE-V), which uses visual representations for routing\footnote{Our implementation builds upon the \href{https://github.com/zjunlp/EasyEdit/tree/main}{EasyEdit} framework.}.

We consider two editing scenarios.
Single edits evaluate whether methods can fix individual recognition failures, while
sequential edits are more practical editing use cases that evaluate whether a model can acquire new concepts over time.
Sequential edits are applied cumulatively in a random order, shared across all methods to ensure comparability.
IKE and MEND are excluded from sequential evaluation: IKE does not modify model parameters and has no mechanism for accumulating edits, while MEND's hypernetwork is trained against the model's initial weights, and, as shown by \citet{lin2024navigating}, its predictions become unreliable under sequential application.
After editing $N$ concepts sequentially, we report average metrics across all $N$ edited concepts.
Results are reported after  $N = \{10, 50, 100, 150, 200, 250\}$ edits. Experiments are repeated for five orders.

Edit sets are derived from each model's recognition failures and therefore cover different concept subsets per model.
These sets largely overlap (90.3--94.3\% pairwise IoU), but we base our conclusions on within-model comparisons across methods rather than comparing a method across models.
Additionally, as each method targets a different component, parameter budgets vary across methods and models (\Cref{tab:edit-config}). Results should, therefore, be read as comparisons between editing strategies under their standard configurations, rather than between updates of equal capacity.

\section{Results}

\subsection{Single Edits}
\begin{figure*}[t!]
\centering
\begin{subfigure}[t]{0.333\textwidth}       \includegraphics[width=\textwidth]{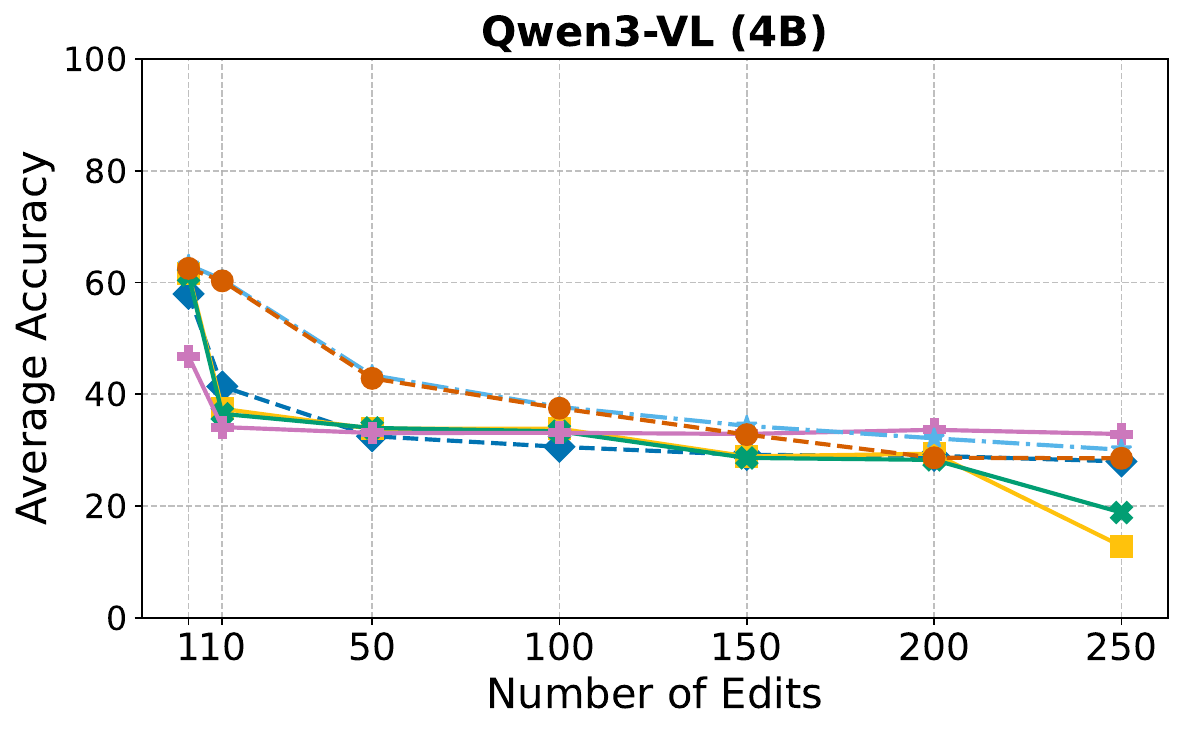}
\end{subfigure}%
\begin{subfigure}[t]{0.333\textwidth}
       \includegraphics[width=\textwidth]{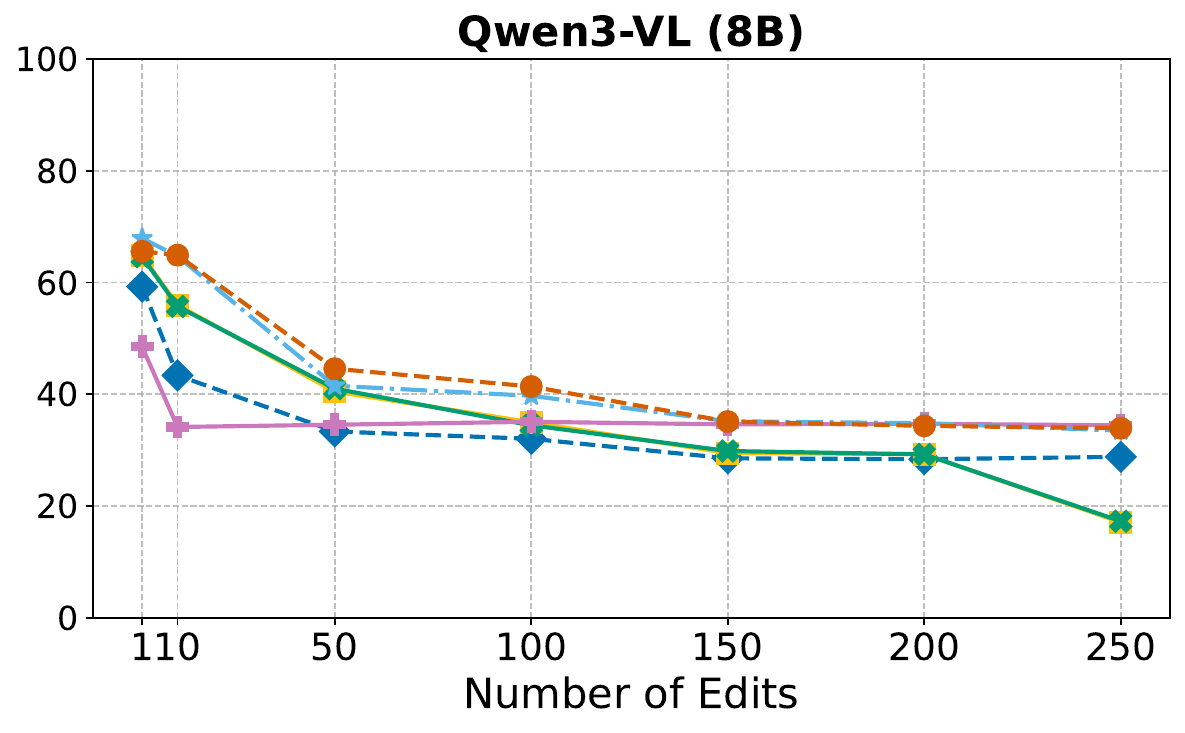}
\end{subfigure}%
\begin{subfigure}[t]{0.333\textwidth}       \includegraphics[width=\textwidth]{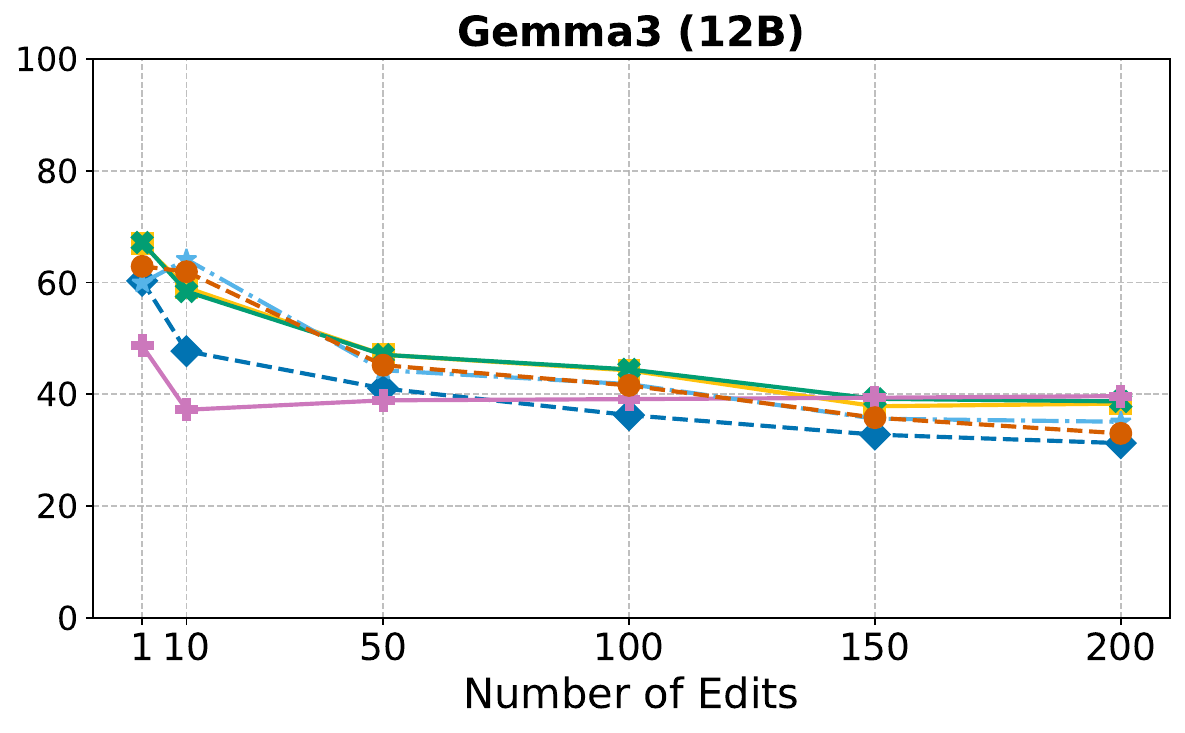}
\end{subfigure}%
\\
\includegraphics[width=0.5\textwidth]{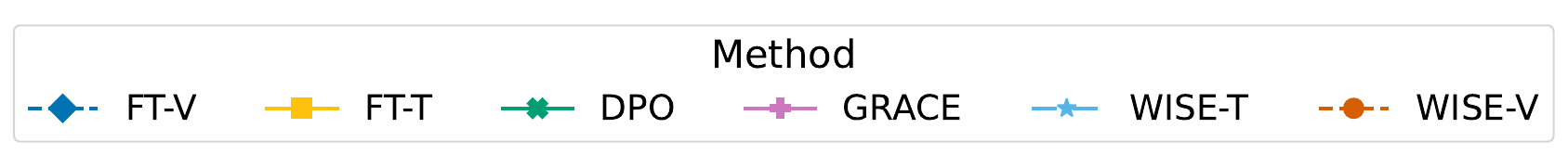}
\caption{Average accuracy of editing metrics as a function of the number of sequential edits.}
\label{fig:sequential_editing_results}
\end{figure*}

As shown in \Cref{tab:main_qwen_4}, all methods achieve successful single edits with high reliability, confirming that fixing individual recognition errors is feasible across architectures.
However, reliability does not guarantee generalization: methods consistently trade off generality and locality, with this tension most pronounced for image-based metrics.
Among post-training methods, finetuning the visual connector (FT-V) negatively impacts both image generality and image locality, suggesting that direct updates to the visual connector disrupt cross-modal alignment rather than supporting targeted concept acquisition.
Finetuning the language model (FT-T) improves image generality substantially, and DPO yields comparable results.

Comparing editing methods, MEND achieves the most consistent performance across models.
WISE also achieves competitive performance, though with a consistent weakness in image locality. WISE-T, which uses text-based locality samples to learn routing, achieves stronger text generality and text locality, while WISE-V, which uses image-based locality samples, recovers image locality at the cost of slightly lower generality.
Together, MEND and WISE represent the most viable approaches for single-concept editing, though neither achieves strong portability.

IKE and GRACE demonstrate opposite failure modes.
IKE achieves near-perfect generality by retrieving demonstration examples at inference time, but this comes at a severe cost to locality.
This inability to leverage in-context text-based demonstration is in line with the findings of \citet{nikandrou2024crope}.
GRACE is at the opposite extreme with minimal generality and near-perfect locality.
This is a consequence of its activation-matching mechanism, as stored edits are retrieved only when input activations fall within a narrow neighborhood, preventing generalization across instances.

Text and image portability are uniformly low across all methods and models, ranging from 7 to 28\%, and do not meaningfully differentiate method families. Even when edits are successful on other measures, the new associations between images and concept labels do not propagate to the model's broader parametric knowledge.
To better understand model behavior, we examine randomly sampled outputs before and after editing. We classify each outcome as \textit{corrected} (prediction changed from incorrect to correct), \textit{damaged} (prediction changed from correct to incorrect), or \textit{unchanged} (prediction matched the pre-edit answer, regardless of correctness).
\Cref{tab:portability_delta} reports the distribution of these outcomes. The overwhelming majority of questions fall in the unchanged category, showing that editing does not meaningfully alter model behavior on portability. Only 12-23\% of questions show any change and among those, corrected and damaged cases occur at comparable rates, yielding near-zero net improvement. These suggest that the low portability scores are not because of errors introduced by editing but reflect pre-existing gaps consistent with the results in \Cref{sec:cross_modal}.

One possible explanation for the low image portability comes from mechanistic work on factual recall from visual inputs \citep{cohen-etal-2025-performance, venhoff2025too_late}: even when recognition succeeds, recall can fail when visual entity information becomes available at deeper model layers than those supporting factual recall.
Most examined methods edit the second half of the language model (\Cref{tab:edit-config}), so an edit may fix recognition while leaving the concept inaccessible at the depth of factual recall.
FT-V, the only method that edits a component before the language model, improves image portability over FT-T only slightly (2--5 points).\footnote{Text portability does not involve visual entity resolution and is therefore not explained by this mechanism.}

\begin{table}[tb]
    \centering
\resizebox{\linewidth}{!}{

    \begin{tabular}{llccc}
\toprule
Model & Portability & Corrected & Damaged & Unchanged \\
\midrule
\multirow{2}{5em}{Qwen3-VL (4B)}  & Text  & 48 & 61  & 367 \\
 & Image & 23 & 29 & 396 \\
\addlinespace
\multirow{2}{5em}{Qwen3-VL (8B)}  & Text  & 47 & 28 & 398 \\
 & Image & 35 & 21 & 366 \\
\addlinespace
\multirow{2}{5em}{Gemma 3 (12B)} & Text  & 43 & 34 & 393 \\
 & Image & 40 & 25 & 370 \\
\bottomrule
    \end{tabular}}
    \caption{Portability outcomes before and after editing.}
    \label{tab:portability_delta}
\end{table}

\subsection{Sequential Edits}

\Cref{fig:sequential_editing_results} shows average accuracy for sequential editing across methods and models. All methods exhibit a decline as the number of edits increases, with performance dropping sharply even at N=50.
Between N=50 and N=200, methods cluster within a narrow accuracy range, with no single approach maintaining a consistent advantage.
However, at N=250, FT-T and DPO collapse to near-zero for both Qwen3-VL variants, reflecting the compounding effect of sequential finetuning of the last language model layer.
GRACE and WISE-T appear more stable at scale, but per-metric results (\Cref{app:sequential_edit}) show that this reflects preserved locality rather than editing success, as reliability and generality decline for both methods.

\paragraph{Forgetting vs.\ Retention}
To separate forgetting of earlier edits from failure to acquire edits in the first place, we evaluate the final Qwen3-VL (4B) checkpoint and compare performance on the earliest and last 25\% of edits (\Cref{tab:early_late}).
For parametric methods (FT-V, FT-T, DPO), recent edits score considerably higher than early ones, indicating forgetting.
WISE shows no such recency effect: performance is similarly low on early and recent edits, including reliability. Its degradation therefore does not stem from forgetting, but is consistent with interference among accumulated edits.

\begin{table}[t]
\centering
\small
\setlength{\tabcolsep}{3pt}
\begin{tabular}{lccc}
\toprule
Method & Reliability & T-Gen & I-Gen \\
\midrule
FT-V   &  8.8 $\rightarrow$ 53.6 &  6.7 $\rightarrow$ 43.2 &  2.7 $\rightarrow$ 5.6 \\
FT-T   & 19.2 $\rightarrow$ 54.4 & 16.0 $\rightarrow$ 39.2 & 10.4 $\rightarrow$ 35.2 \\
DPO    & 16.8 $\rightarrow$ 56.8 & 16.8 $\rightarrow$ 44.0 & 15.3 $\rightarrow$ 34.7 \\
WISE-T & 29.6 $\rightarrow$ 24.8 & 28.0 $\rightarrow$ 26.0 & 20.8 $\rightarrow$ 25.6 \\
WISE-V & 29.6 $\rightarrow$ 31.2 & 32.0 $\rightarrow$ 28.7 & 26.4 $\rightarrow$ 28.0 \\
\bottomrule
\end{tabular}
\caption{Accuracy of Qwen3-VL (4B) after sequential editing, on the earliest $\rightarrow$ last 25\% of edits.}
\label{tab:early_late}
\end{table}

\section{Conclusion}
We introduce CulturalBind, a multidimensional benchmark for assessing textual and visual knowledge gaps of cultural concepts.
Our cross-modal evaluation reveals that VLMs demonstrate stronger parametric knowledge in text-only settings than when identifying the same concepts from images, a gap that widens for visually-grounded QA, which requires factual knowledge recall through visual referents.
Additionally, even when conditioning on concepts a model can describe, visual recognition remains weak, indicating that textual and visual knowledge are largely decoupled.

We then use CulturalBind as an editing benchmark targeting VLM recognition failures.
While editing methods can reliably correct individual recognition errors without disrupting unrelated model knowledge, portability remains uniformly low. This indicates that edited associations do not propagate to the model's broader parametric knowledge of the concept.
Under sequential editing, performance degrades sharply: parametric methods forget earlier edits, while memory-based methods preserve locality but fail to reliably recall even recent edits.
Overall, our findings point to cross-modal consistency as a key missing objective, both during pretraining and in post-hoc adaptation.
We hope CulturalBind motivates future work on methods that consolidate multimodal knowledge rather than patch isolated associations, moving towards VLMs that are adaptable across the diversity of global cultural concepts.

\section{Limitations}
First, experiments are conducted exclusively in English. Inclusive cultural evaluation ultimately requires multilingual assessment, as the visual-linguistic binding gap measured in English may not reflect the experience of speakers from the concept's origin language.
At the same time, known performance disparities across languages in VLMs \citep{yue2024pangea,salazar2025kaleidoscope} make it difficult to disentangle gaps attributable to cultural concept coverage from those attributable to language-level representation more broadly.
Additionally, we acknowledge the limitations of our region-based operationalization of culture, following the geographic metadata available in source datasets, and the uneven cultural coverage in \dataset with South American and African sub-regions underrepresented, reflecting known inequalities in Wikidata and Wikipedia coverage \citep{10.3389/fphy.2018.00054}.
Finally, the evaluated editing methods represent a selection of the current landscape that spans varying editing mechanisms and target layers across the visual projector and language model layers.
However, these methods do not explicitly target cross-modal integration, which remains an important research direction for future work.

\section{Ethical Considerations}
This work diagnoses failures in the recognition of cultural concepts and the integration of knowledge in VLMs, with the goal of motivating more effective adaptation methods.
The concepts evaluated represent cultural artefacts and practices, and absent or inaccurate model knowledge can disadvantage users from underrepresented groups \citep{prabhakaran2022cultural}.
Our evaluation anchors cultural knowledge to English Wikipedia, whose documentation of cultural knowledge is itself shaped by WEIRD biases. While performance on our benchmark should not be interpreted as a comprehensive measure of cultural competence, it allows us to separate between concepts a model genuinely lacks knowledge of and those it fails to access through visual referents.
We hope this work contributes to advancing the cultural adaptability and multimodal grounding of VLMs.

\section*{Acknowledgements}
This work was supported by the Edinburgh International Data Facility (EIDF) and the Data-Driven Innovation Program at the University of Edinburgh.
In addition, the authors acknowledge the use of the HWU high-performance computing facility (DMOG) and associated support services in the completion of this work.

\newpage

\bibliography{custom}
\clearpage

\appendix

\begin{figure*}[!th]
    \centering
\begin{subfigure}[t]{0.50\textwidth}
\includegraphics[width=\textwidth]{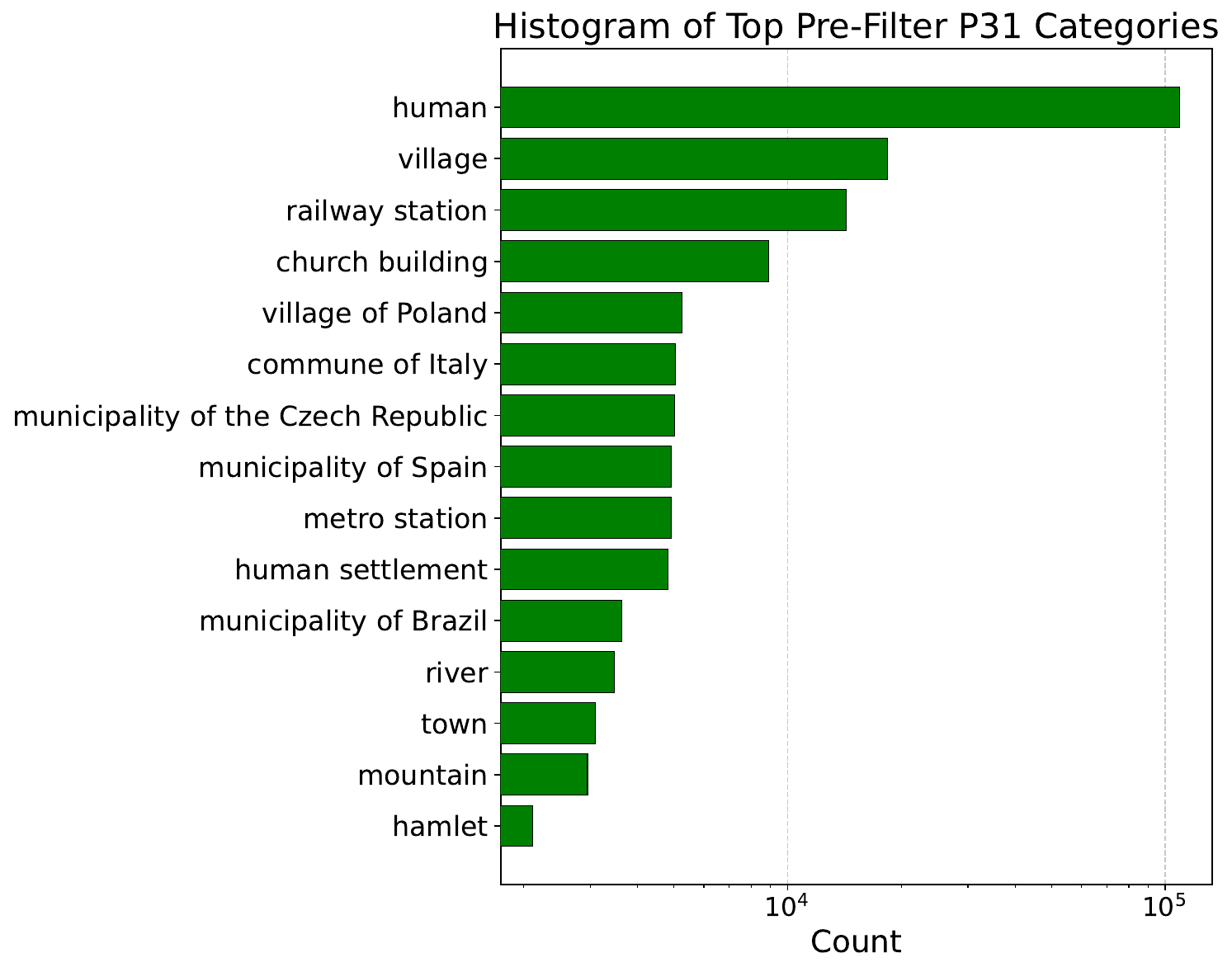}
\end{subfigure}%
\begin{subfigure}[t]{0.50\textwidth}
\includegraphics[width=\textwidth]{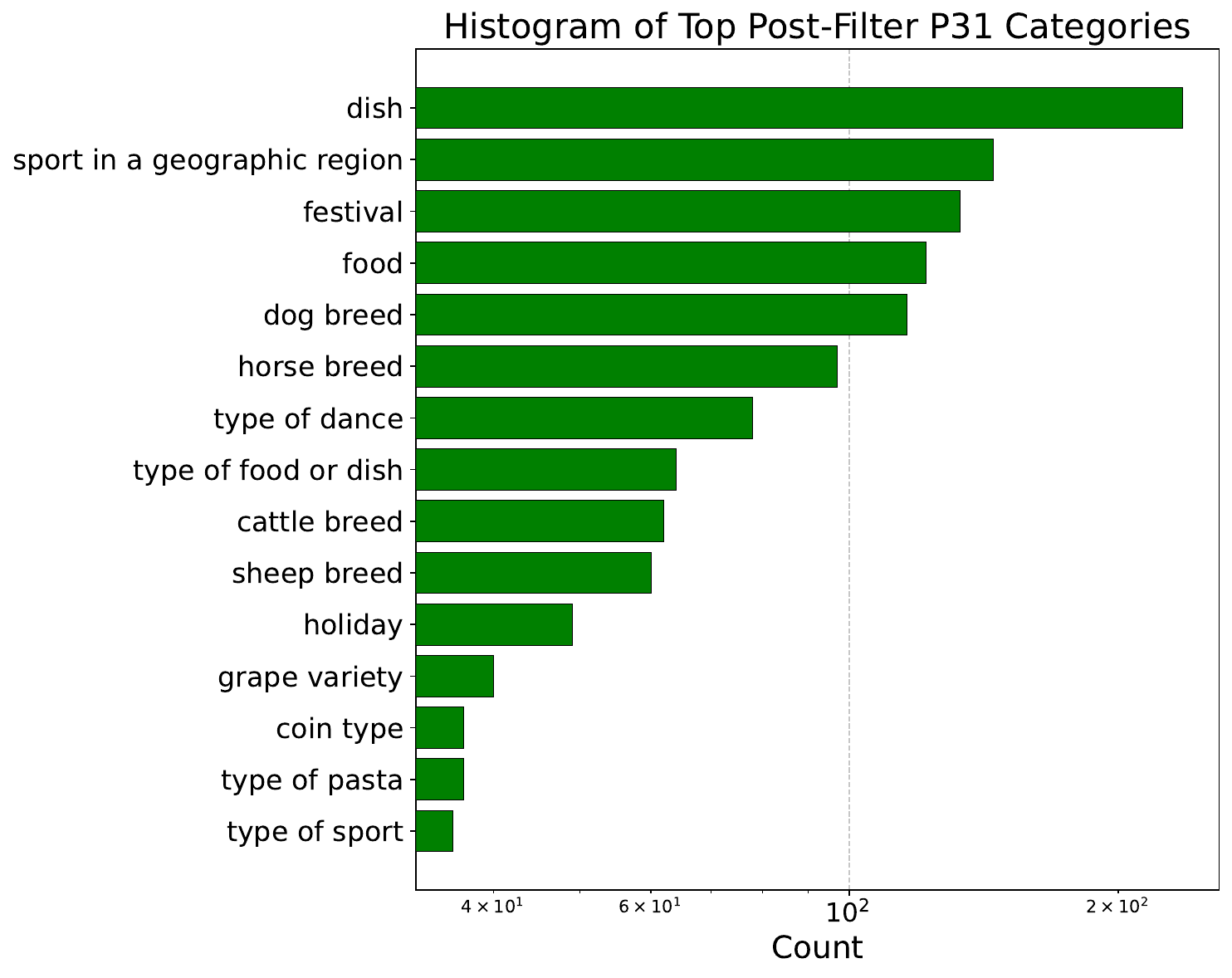}
\end{subfigure}
    \caption{Distribution of the 15 largest Wikidata \textit{instance of} (P31) groups among CulturalGround entities with English Wikipedia pages and multiple images. Before filtering, the majority of items belong to named-entity categories, which do not satisfy the concept criterion and are excluded at the group level.}
    \label{fig:cultural_ground_hist}
\end{figure*}

\begin{figure*}
    \centering
\includegraphics[width=\linewidth]{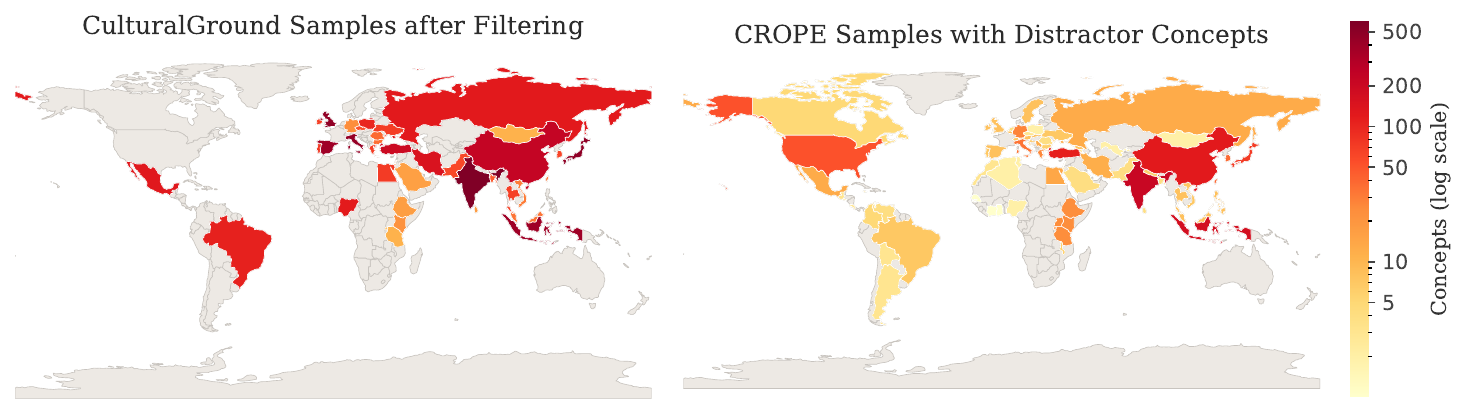}
    \caption{Geographic distribution of concepts in CulturalGround (after filtering, left) and CROPE (after retrieving images for distractor concepts, right).}
    \label{fig:geo_map}
\end{figure*}

\section{Dataset Details}\label{appendix:edit-data}
\subsection{Data Sources}

CulturalGround \citep{nyandwi-etal-2025-grounding} is a multilingual dataset of 22M culturally relevant VQA questions covering 42 countries and 39 languages, with cultural entities and images sourced from Wikidata.
For the purpose of this study, it provides an initial pool of culturally relevant entities.
However, the majority are named entities:
as shown in Figure~\ref{fig:cultural_ground_hist}, the distribution of Wikidata \textit{instance of} (P31) types is dominated by
individual people (109K items), followed by geographic administrative units and infrastructure, none of which correspond to generalizable visual categories.
We therefore filter at the group level, retaining only P31 types whose label denotes a concrete visual concept category (e.g., \textit{type of musical instrument}) and excluding proper-noun categories such as \textit{human}, \textit{business}, or \textit{railway station}.
We further restrict to items that link to English Wikipedia pages, from which background knowledge can be extracted.

Note that while CulturalGround provides a VQA dataset, we follow a different QA generation approach suited to our evaluation objectives. CulturalGround prioritises multilinguality and broad coverage of cultural entities, providing on average 1.5 encyclopedic questions per entity.
Our QA generation instead targets concept-level knowledge evaluation, generating on average 4.86 encyclopedic and 4.88 visual attribute questions per
concept.

CROPE \citep{nikandrou2024crope} is a dataset for probing in-context adaptation to culture-specific concepts,  building on top of existing multimodal reasoning work, MaRVL \citep{liu-etal-2021-visually} and MCC \citep{li-zhang-2023-cultural}.
By definition, the provided concepts satisfy the concept criterion, but images cover only 340 concepts from five language origins (Indonesian, Swahili, Tamil, Turkish, and Mandarin Chinese).
CROPE frames concept recognition as a binary question, pairing each target concept with an adversarially selected distractor, visually or functionally similar, to test fine-grained discrimination. We consider both positive and negative concepts, retrieving additional images for each distractor.
The region mapping is completed by querying Wikidata for \textsc{country of origin} (P495), \textsc{country} (P17), and \textsc{location} (P276).

\subsection{Human Validation}\label{app:qa_audit}
All Openverse and CulturalGround images retained after automated filtering were manually inspected by the authors, removing 96 images and yielding the final 3,216. CROPE images had already been validated by native-speaker annotators \citep{nikandrou2024crope}.
For the generated QA pairs, we sample 20 pairs per question category (200 in total, see \Cref{tab:background_questions}) and annotate each pair on two criteria: (i) \textit{question clarity}, whether the question is well-formed with a single correct answer, and (ii) \textit{answer faithfulness}, whether the answer is supported by the source Wikipedia text.

\begin{table}[h]
\centering
\small
\begin{tabular}{lcc}
\toprule
 & \makecell{Question\\Clarity} & \makecell{Answer\\Faithfulness} \\
\midrule
Encyclopedic QA     & 89.3 & 94.6 \\
Visual Attribute QA & 90.5 & 96.5 \\
\bottomrule
\end{tabular}
\caption{Manual audit of 200 generated QA pairs (\%).}
\label{tab:qa_audit}
\end{table}

As shown in \Cref{tab:qa_audit}, all audited QA pairs were grammatically correct and coherent.
Question clarity errors mostly result from underspecified questions (e.g., ``Where'' questions that do not specify whether the answer should be a region, country, or state), while faithfulness errors mostly result from partial answers (e.g., the reference answer includes only ``coconut'' instead of ``lentils and coconut'').

\subsection{Dataset Construction Prompts}\label{appendix:prompt}

Below, we report the prompts for 1) filtering CulturalGround for concept selection (\Cref{fig:image_culturalground_filter_prompt}), 2) filtering images (\Cref{fig:image_filter_prompt}), 3) extracting visual descriptions (\Cref{fig:visual_desc_prompt}), 4) extracting encyclopedic question-answer pairs (\Cref{fig:encyclopedic_prompt}).

\begin{figure*}[ht]
    \centering
    \scriptsize
    \begin{tcolorbox}[title=Prompt for CulturalGround Concept Filtering]

\vspace{-1mm}
\#\# Task
\vspace{1mm}

Your task is to filter Wikidata "instance of" (P31) groupings for a Visual Cultural Concept Learning dataset.

\vspace{3mm}
\#\# Context
\vspace{1mm}

You are given Wikidata items grouped by their P31 property. You must identify P31 groupings where the individual items represent concrete, concept categories rather than specific entities.

\vspace{1mm}
\#\# Instructions
\vspace{1mm}

Classify a Wikidata P31 grouping based on its label and sample items.

\textbf{Keep} groupings if the items are:

1. Concept categories (common nouns): Each item defines a type or category with multiple real-world instances (e.g., "Golden Retriever" is a category containing many individual dogs).

2. Visually distinguishable: Each item has distinct, recognizable visual features. This includes physical objects with shared visual attributes (foods, instruments, clothing, animal breeds), activities or performances with visible form, events or rituals with recognizable settings, costumes, or specific ceremonial objects.

\textbf{Remove} groupings if the items are:

1. Proper Nouns: Each item is a unique instance or named entity, such as specific locations, organizations, or people.

2. Abstract or non-visual: The items represent abstract concepts, roles, or categories that lack a consistent visual representation. (e.g., languages, specific competitions, administrative divisions, business types)

\vspace{1mm}
\#\# Examples
\vspace{1mm}

Instance of: "animal breed"\\
Sample entities: Golden Retriever, Siamese Cat, Clydesdale Horse\\
Output: \{"reason": "These are categories of animals with shared visual features and many real-world instances.", "decision": "keep"\}

Instance of: "record label"\\
Sample entities: Motown, Deutsche Grammophon, Sub Pop\\
Output: \{"reason": "These are specific corporate entities, not a visual object category.", "decision": "remove"\}

Instance of: "railway station"\\
Sample entities: Joachimsthal Kaiserbahnhof, Hoffenheim station, Büsum station\\
Output: \{"reason": "These are specific named railway stations, not types of stations.", "decision": "remove"\}

Instance of: "type of musical instrument"  \\
Sample entities: Sitar, Djembe, Erhu\\
Output: \{"reason": "These are categories of musical instruments with multiple real-world instances.", "decision": "keep"\}

Instance of: "film festival"\\
Sample entities: Edinburgh International Film Festival, Prague Independent Film Festival, Cannes\\
Output: \{"reason": "These are specific, named events tied to a location and time, and lack visually distinguishing features between them.", "decision": "remove"\}

Instance of: "festival"\\
Sample entities: Diwali, Glastonbury, Carnival of Venice\\
Output: \{"reason": "These are specific, recurring named events, with characteristic visual elements (decorations, costumes, settings).", "decision": "keep"\}

\vspace{1mm}
\#\# Output Format
\vspace{1mm}

Respond with a JSON object containing the following fields:

\{"reason": "<short reasoning>", "decision": "keep" | "remove"\}
\end{tcolorbox}
\caption{Prompt for filtering the CulturalGround entities.}
\label{fig:image_culturalground_filter_prompt}
\end{figure*}

\begin{figure*}[!h]
    \centering
    \small
    \begin{tcolorbox}[title=Prompt for Image Filtering]

\vspace{-1mm}
\#\# Task

Determine whether the image provides sufficient visual evidence to confidently answer the recognition question using the image alone.
You must respond with either "keep" or "remove".

\vspace{2mm}
\#\#\# Decision Rule

Could a knowledgeable person correctly answer the question from the visual appearance alone, without relying on any text in the image?

\vspace{2mm}
\#\# Instructions

\textbf{Keep} if ALL the following criteria are met:
\vspace{-1mm}
\begin{itemize}[leftmargin=*]
\itemsep0em
\item The image shows a clear example of the concept
\item The visual appearance is prototypical and visually distinctive enough to support confident recognition
\item The concept can be identified from visual features alone (no need to read any text)
\end{itemize}
\vspace{1mm}

\textbf{Remove} if ANY of the following criteria are met:
\vspace{-1mm}
\begin{itemize}[leftmargin=*]
\itemsep0em
\item The concept cannot be confidently identified from visual features alone
\item The example is atypical, ambiguous, or heavily occluded
\item Recognition depends on reading text present in the image (labels, captions, logos, signage)
\end{itemize}

\vspace{2mm}
\#\#  Output Format

Respond with a JSON object containing the following fields:

\{"reason": "<short reasoning>", "decision": "keep" | "remove"\}

\vspace{2mm}
\#\# Context

\textbf{Concept:} \{concept\}

\textbf{Wikipedia Summary:} \{summary\}

\textbf{Question:} \{question\}

\textbf{Image:} \{image\}

\end{tcolorbox}
\caption{Prompt for filtering the dataset images.}
\label{fig:image_filter_prompt}
\end{figure*}

\begin{figure*}[ht]
    \centering
    \small
    \begin{tcolorbox}[title=Prompt for Visual Description Extraction]

\vspace{1mm}
\#\# Task
\vspace{1mm}

From the following Wikipedia text about \textbf{\{concept\}}, extract ONLY the visual descriptions explicitly stated in the text that could help a vision-language model identify \{concept\} in images.

\vspace{3mm}
\#\# Instructions
\vspace{1mm}

\begin{enumerate}[leftmargin=*]
\item Extract only visual features that are directly stated in the text. Focus only on features that would be observable in an image:
\begin{itemize}
    \item What would you see if looking at this concept?
    \item What visual details would help distinguish it from similar concepts?
\end{itemize}

\item Do not add interpretations, inferences, or outside knowledge.
\item Do not include any references to cultural or regional origins.
\item Do not include historical information that no longer applies to the current form of the concept (e.g., outdated definitions, or previous uses).
\item Include frequency qualifiers when present (e.g., "typically," "rarely").
\end{enumerate}

\vspace{3mm}
\#\# Examples
\vspace{1mm}

\{examples\}

\vspace{3mm}
\#\# Output Format
\vspace{1mm}

Return the visual descriptions into a JSON object with the following structure, categorizing each description into one of the following mutually exclusive categories:

\vspace{-1mm}
\begin{itemize}[leftmargin=*]
\itemsep0em
\item Type of (e.g., dish, beverage, animal species, dance, musical genre, monument, ceremony, tool)
\item Physical attributes (e.g., colors, shapes, materials, sizes, textures, composition, clothing)
\item Actions (e.g., gestures, movement patterns, and other observable actions that characterize how the concept is expressed or performed. Do not refer to general knowledge, such as preparation, animal behaviors, or tool effects.)
\item Contextual cues (e.g., typical settings, backgrounds, common co-occurring objects)
\end{itemize}

If a category has no relevant visual descriptions, return an empty list for that category.
\end{tcolorbox}
\caption{Prompt for extracting visual descriptions from the Wikipedia page.}
\label{fig:visual_desc_prompt}
\end{figure*}

\begin{figure*}[ht]
    \centering
\small
    \begin{tcolorbox}[title=Prompt for Encyclopedic Knowledge QA Generation]

You are a specialized QA generator creating a dataset for closed-book knowledge testing.

\vspace{1mm}
\#\# Task
\vspace{1mm}

Given a Wikipedia extract and visual descriptions, generate up to 5 QA pairs testing for background knowledge that is NOT present in the visual description and cannot be inferred by looking at an image of the concept.

\vspace{1mm}
\#\# Instructions
\vspace{1mm}

\begin{enumerate}
    \item REFERENCE: All questions must name the concept explicitly.
    \item CLARITY: Questions must be unambiguous, with a single correct answer. Avoid questions that could have multiple valid answers or require subjective interpretation.
    \item NO VISUALS: If a person could see the answer in a photograph, do not ask the question.
    \item SOURCE GROUNDING: Answers must be found within the provided 'EXTRACT'.
    \item BREVITY: Answers must be a single word or a very short phrase (max 5 words).
    \item CATEGORIZATION: Every question must map to one of the specific categories below.
\end{enumerate}

\vspace{1mm}
\#\# Question Categories
\vspace{1mm}

\begin{enumerate}
    \item cultural association (ethnic groups, religions, traditions, festivals, cultural practices)
    \item historical (dates, events, historical figures)
    \item geographic (geographic or regional origins, locations)
    \item function (purpose, use case, or role)
    \item composition (ingredients, preparation methods, or components that are not visually identifiable)
    \item linguistic (etymology, name meaning, language of origin)
    \item comparison (equivalents or related items)
    \item other (any other non-visual background knowledge)
\end{enumerate}

\vspace{1mm}
\#\# Output Format
\vspace{1mm}

Output only a  valid JSON object:

{{
    "questions": [
        {{"question": "...", "answer": "...", "category": "cultural association" | "historical" | "geographic" | "function" | "composition" | "linguistic" | "comparison" | "other"}}
    ]
}}

\vspace{3mm}
\#\# Examples
\vspace{1mm}

\{examples\}

    \end{tcolorbox}
\caption{LLM prompt to generate encyclopedic questions.}
\label{fig:encyclopedic_prompt}
\end{figure*}

\subsection{Additional Dataset Statistics}

\begin{table}[]
\centering
\small\setlength{\tabcolsep}{2pt}
\begin{tabular}[t]{lrr}
\toprule
Domain & Concepts & Samples \\
\midrule
Foods \& Beverages       & 432 & 1{,}235 \\
Animals \& Plants        & 177 &   508 \\
Music \& Performing Arts & 152 &   543 \\
Artefacts                &  135 & 462 \\
Traditions               & 126 &   300 \\
Sports                   &  69 &   168 \\
\midrule
Total                    & 1{,}090 & 3{,}216 \\
\bottomrule
\end{tabular}
\caption{\dataset distribution by domain.}\label{tab:dataset_domain}
\end{table}

\begin{table}
  \centering
    \addtolength{\tabcolsep}{-0.3em}
\small
\begin{tabular}{llc}
    \toprule
 \textbf{Question} & \textbf{Description} & \textbf{Count}\\
\midrule
Historical & Dates, events, historical figures & 1,322 \\
Geographic & Regional origins, specific locations & 1,163 \\
Linguistic & Etymology, meaning, language of origin & 868 \\
Cultural & Ethnic groups, practices, traditions & 633 \\
Function & Purpose, use case, role & 449 \\
Comparison & Similarities, equivalents, related items & 441 \\
Composition & Non-visible ingredients or components & 372\\
\midrule
Physical & Shape, color, size, visible components & 3,043\\
Contextual & Settings, accompanying objects & 1,341 \\
Action & Movements, gestures, interactions & 937 \\
\bottomrule
    \end{tabular}
\caption{Distribution of 5,298 encyclopedic (top) and 5,321 visual attribute (bottom) knowledge questions.}
\label{tab:background_questions}

\end{table}

Table~\ref{tab:dataset_domain} reports the distribution of concepts and images across the six cultural domains in \dataset. Foods \& Beverages is the largest domain, accounting for nearly 40\% of concepts, while Sports is the smallest, reflecting the filtering criterion that excludes specific competitions and  athletic events. The remaining domains are broadly balanced across concrete object categories (Artefacts, Animals \& Plants) and performative or ceremonial human activities (Music \& Performing Arts, Traditions).

Table~\ref{tab:background_questions} reports the distribution of knowledge questions across encyclopedic and visual attribute categories.
Among encyclopedic questions, Historical, Geographic, and Linguistic categories together account for over 60\% of questions.
Cultural and Functional questions provide complementary coverage of context and use.
Following in terms of number, Composition and Comparison questions, probe knowledge that is particularly resistant to visual inference, such as non-visible ingredients or cross-cultural equivalents.
Among visual attribute questions, Physical characteristics are the most frequent, followed by Contextual and Action questions, which capture relational and motion-related visual features.
Together, the ten question categories span a diverse set of knowledge types.

\subsection{Dataset Editing Examples}

\begin{figure*}[h]
    \includegraphics[width=\textwidth]{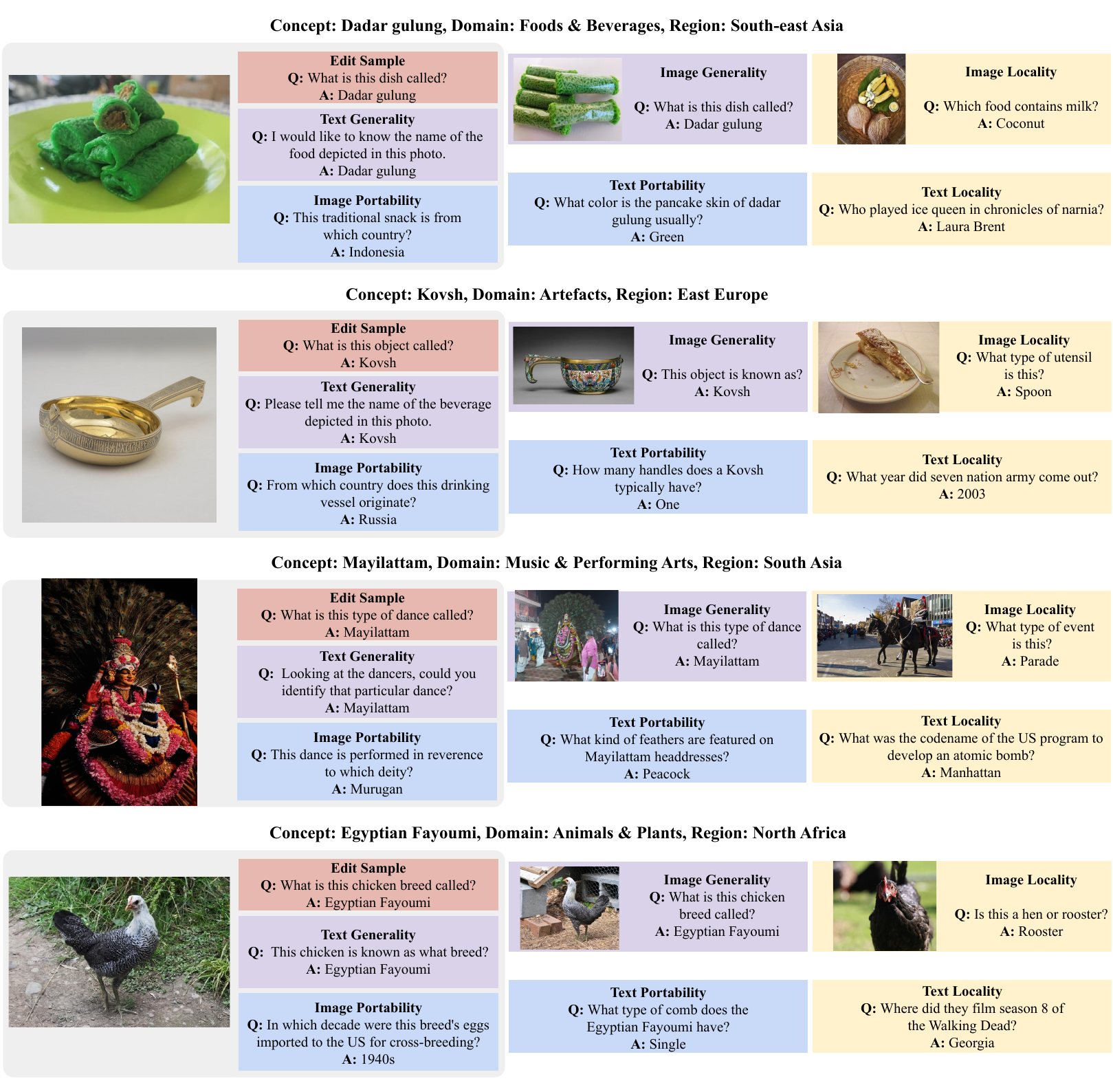}
        \caption{\dataset editing examples.}
    \label{fig:cced_examples}
\end{figure*}

Figure~\ref{fig:cced_examples} illustrates representative editing examples from \dataset across four cultural domains and regions, showing the edit sample and its associated generality, locality, and portability probes.

\subsection{Concept-Level vs.\ Entity-Level Edits}\label{app:entity_vs_concept}
We compute the SigLIP2 \citep{tschannen2025siglip} cosine similarity between each edit image and its paired generality and locality images in \dataset and two entity-based editing benchmarks, VLKEB \citep{huang2024vlkeb} and the entity editing subset of MMKE-Bench \citep{dummke}, using 2,000 samples each.
As shown in \Cref{tab:entity_vs_concept}, \dataset edit images are less visually similar to their generality images (mean 0.756) than those of VLKEB (0.853; Mann-Whitney U test $p<10^{-5}$, Cliff's $\delta=-0.48$) and MMKE-Bench (0.845; $p<10^{-5}$, $\delta=-0.44$), while remaining more similar than the locality images.
This supports the hypothesis that generalizing across concept instances involves greater visual variation than identifying the same entity under different conditions.

\begin{table}[h]
\centering
\small
\setlength{\tabcolsep}{3pt}
\begin{tabular}{lcc}
\toprule
 & Generality & Locality \\
\midrule
\dataset     & 0.756 $\pm$ 0.124 & 0.548 $\pm$ 0.095 \\
VLKEB        & 0.853 $\pm$ 0.073 & 0.688 $\pm$ 0.113 \\
MMKE-Bench   & 0.845 $\pm$ 0.077 & 0.425 $\pm$ 0.084 \\
\bottomrule
\end{tabular}
\caption{SigLIP2 cosine similarity (mean $\pm$ sd) between edit images and their paired generality and locality images.}
\label{tab:entity_vs_concept}
\end{table}

\subsection{License}
\dataset is released publicly upon publication under CC BY-SA 4.0.
Images are sourced from Wikimedia Commons and Openverse under CC0, CC BY, and CC BY-SA licenses.
Wikipedia text used for knowledge QA generation is licensed under CC BY-SA 4.0.

\section{LLM-Judge Evaluation}\label{appendix:llm-judge}

\begin{figure*}
    \centering
    \begin{tcolorbox}[title=Open-Ended Concept Description Evaluation Prompt]

\vspace{1mm}
\#\# Task

You are a cultural knowledge expert. Given a question asking a model to describe a cultural concept, a reference Wikipedia summary, and the model's prediction, determine if the model demonstrates knowledge of the concept.

\vspace{1mm}
\#\# Goal
\vspace{1mm}

Decide if the model is correctly identifying the SAME concept described in the reference Wikipedia summary.

\vspace{1mm}
\#\# Scoring Criteria
\vspace{1mm}

\begin{itemize}
    \item \textbf{Score 1}: The model's prediction clearly refers to the same cultural concept as the Wikipedia summary. Minor omissions, factual errors, or additional true details are acceptable, as long as the prediction captures the key definition of the concept.
    \item \textbf{Score 0}: The model fails to demonstrate knowledge of the concept. This includes:
    \begin{itemize}
        \item Hallucinating a completely different definition.
        \item Describing a different concept with the same name.
        \item Refusing to answer or claiming not to know the concept.
        \item Providing generic text that does not specifically describe the concept.
    \end{itemize}
\end{itemize}

\vspace{1mm}
\#\# Output Format
\vspace{1mm}

Respond only with the following JSON structure and nothing else:

\{
  "explanation": "<brief explanation for the score>",
  "score": <0 | 1>
\}
    \end{tcolorbox}
\caption{Evaluation rubric for Open-Ended Description.}
\label{fig:evaluation_rubric_description}
\end{figure*}

Below, we report the prompts used for the evaluation of Open-Ended Concept Description (\Cref{fig:evaluation_rubric_description}), factual QA tasks (Encyclopedic, Attribute, and Visually-Grounded Encyclopedic QA, \Cref{fig:evaluation_rubric_factual}), and Recognition QA (\Cref{fig:evaluation_rubric_recongition}).

To validate the reliability of automated evaluation, we independently annotated N=150 samples blind to the judge scores, stratified by score across all three evaluation dimensions: Open-ended Description (N=40), Text-based Factual QA (60 samples), and Image-based Recognition QA (50 samples). We use Qwen3-32B \citep{yang2025qwen3} as the LLM judge, as it is openly available and results are therefore fully reproducible. We compute Quadratic Weighted Kappa (QWK) between our annotations and the judge scores, which penalizes larger ordinal discrepancies more heavily. Agreement is strong across all dimensions with an overall QWK of 0.856.
These results support the use of Qwen3-32B as a reliable proxy for reference-based evaluation.

\begin{figure*}[ht]
    \footnotesize
    \centering
    \begin{tcolorbox}[title=Factual QA Evaluation Prompt]

\vspace{-1mm}
\#\# Task

You are given a factual question, a ground-truth reference answer written by experts, and a candidate answer.
Your task is to rate the accuracy of the candidate answer considering both the question and the reference answer.

\#\# Scoring Rubric (0-2)

Use a scale of 0-2:

0 - Incorrect: The candidate answer contradicts the reference, gives an unrelated fact, or only restates information present in the question.

1 - Partially correct: The candidate answer overlaps with the reference answer but is incomplete, under-specific, or only provides one instance of a broader category. The answer must add meaningful information beyond what can be directly inferred from the question itself and not contradict the reference.

2 - Correct: The candidate answer is fully consistent to the reference answer. It expresses the same fact, an equivalent paraphrase, or a standard instance of the same concept. No essential information is missing.

\#\# Examples

\#\#\# Example 1\\
Question: What is the country of origin for Almdudler?\\
Reference answers: Austria\\
Candidate answer: germany\\
Output: \{
  "explanation": "Clear contradiction. The correct country is Austria, not Germany.",
  "score": 0
\}

\#\#\# Example 2\\
Question: What type of meat is typically used in chanakhi?\\
Reference answers: Lamb Stew\\
Candidate answer: lamb\\
Output: \{
  "explanation": "In the context of this question, 'lamb' is equivalent to 'lamb stew'",
  "score": 2
\}

\#\#\# Example 3\\
Question: Which global regions have been influenced by the spread of chapatis through Indian immigrants?
Reference answers: Central Asia, Southeast Asia, East Africa, and the Caribbean\\
Candidate answer: africa\\
Output: \{
  "explanation": "'Africa' overlaps with 'East Africa' but is under-specific and incomplete.",
  "score": 1
\}

\#\#\# Example 4\\
Question: At which type of events is rendang traditionally served?\\
Reference answers: ceremonial occasions\\
Candidate answer: wedding\\
Output: \{
  "explanation": "Incomplete candidate answer because 'wedding' is only one type of ceremonial occasions.",
  "score": 1
\}

...

\#\# Output Format

Respond only with the following JSON structure and nothing else:

\{
  "explanation": "<short explanation>",
  "score": <0 | 1 | 2>
\}
    \end{tcolorbox}
\caption{Evaluation rubric for factual QA tasks.}
\label{fig:evaluation_rubric_factual}
\end{figure*}

\begin{figure*}[ht]
    \footnotesize
    \centering
    \begin{tcolorbox}[title=Visual Recognition Evaluation Prompt]

\vspace{-1mm}
\#\# Task

You are given a question, a set of gold-standard reference answers written by experts. Your task is to rate the accuracy of the candidate answer considering both the question and the reference answers.

\#\# Scoring Rubric (0-2)

Use a scale of 0-2:

0 - Incorrect: The candidate answer contradicts the references, gives an irrelevant answer, or only restates information present in the question.

1 - Partially correct: The candidate answer overlaps with the reference answers but is incomplete, ambiguous or lacks the specificity of the reference answers.

2 - Correct: The candidate answer matches the reference answer. It provides an equivalent term, or correctly identifies the object at the appropriate level of specificity appropriate to the question.

\#\# EXAMPLES

\#\#\# Example 1 \\
Question: What traditional clothing is the woman wearing?\\
Reference answers: hanbok\\
Candidate answer: kimono\\
Output: \{
    "explanation": "Hanbok is Korean traditional clothing, while kimono is Japanese.",
    "score": 0
\}

\#\#\# Example 2 \\
Question: What is this dish?\\
Reference answers: jiaozi, dumplings, gyoza\\
Candidate answer: gyōza\\
Output: \{
    "explanation": "'Gyōza' is listed among the reference answers.",
    "score": 2
\}

\#\#\# Example 3 \\
Question: What is the rice dish in the image called?\\
Reference answers: pilaf\\
Candidate answer: fried rice\\
Output: \{
    "explanation": "'Pilaf' and 'fried rice' are different dishes.",
    "score": 0
\}

\#\#\# Example 4 \\
Question: What is this flower called?\\
Reference answers: chamomile\\
Candidate answer: daisy\\
Output: \{
    "explanation": "While chamomile belongs to the daisy family, the candidate is less specific than the reference.",
    "score": 1
\}

...

\#\# Output Format

Respond only with the following JSON structure and nothing else:

\{
  "explanation": "<short explanation>",
  "score": <0 | 1 | 2>
\}
    \end{tcolorbox}
\caption{Evaluation rubric for visual recognition QA tasks, adapted from LAVE \citep{manas2024lave}.}
\label{fig:evaluation_rubric_recongition}
\end{figure*}

\section{Implementation Details}

For each model, we construct stratified train, development, and test splits at the concept level with a 55/10/35 ratio, ensuring that all images of a given concept appear in the same split. Edit sets are then derived per model by filtering each split for the concepts the model fails to recognize, and results are reported on the test split. Methods that require additional data draw it from the train split: MEND's hypernetwork is trained on it, and IKE retrieves its in-context demonstrations from it.

We allow up to 30 updates per edit, with early stopping to prevent overfitting to the edit sample.
We follow the default hyperparameters of each method and grid-search over the learning rate and, where applicable, the edit layer, selecting the configuration with the highest average accuracy on single edits on the development split. The selected configurations are reused unchanged in the sequential setting.
For FT-V, FT-T, and DPO, we search over learning rates in $\{1, 3, 5, 8\} \times \{10^{-5}, 10^{-4}\}$. For GRACE and WISE, which apply edits at a single intermediate layer, we additionally sweep over mid-to-late layers.

\Cref{tab:edit-config} details the edited component and parameter counts for each method per model.
Budgets vary with both the method and the model architecture.
This is most visible for FT-V: Gemma 3 reduces image patches with 2D average pooling followed by a linear projection, whereas Qwen3-VL concatenates $2\times2$ patch groups along the feature dimension before a 2-layer MLP, giving its connectors six to nine times as many trainable parameters.

\begin{table}[t]
\centering
\scriptsize
\setlength{\tabcolsep}{2pt}
\renewcommand{\arraystretch}{1.15}
\begin{tabular}{@{}llccc@{}}
\toprule
\textbf{Method} & \makecell[c]{\textbf{Edited}\\\textbf{Component}}  & \makecell[c]{\textbf{Modified}\\\textbf{Params}} & \makecell[c]{\textbf{Added}\\\textbf{Params}} & \makecell[c]{\textbf{Auxiliary}\\\textbf{Component}}  \\
\midrule
\multicolumn{5}{@{}l}{\textit{Qwen3-VL (4B), 36 LM layers}} \\
\midrule
FT-V   & VL connector      & 27M & 0 & --- \\
FT-T   & $W_{down}$, layer 36  & 25M & 0 & --- \\
DPO    & $W_{down}$, layer 36  & 25M & 0 & --- \\
IKE    & ---              & 0   & 0   & SentenceBERT  \\
MEND   & FFN, layer 36 & 75M & 0 & Hypernet. (189M) \\
GRACE  & $W_{down}$ out, layer 28          & 0   & \makecell[r]{key: 9{,}728\\value: 2{,}560} & ---\\
WISE   &  $W_{down}$, layer 22 & 0   & 25M & --- \\
\midrule
\multicolumn{5}{@{}l}{\textit{Qwen3-VL (8B), 36 LM layers}} \\
\midrule
FT-V   & VL connector      & 40M & 0 & --- \\
FT-T   & $W_{down}$, layer 36   & 50M & 0 & --- \\
DPO    & $W_{down}$, layer 36    & 50M & 0 & --- \\
IKE    & ---              & 0   & 0   & SentenceBERT \\
MEND   & FFN, layer 36 & 150M & 0 & Hypernet. (252M) \\
GRACE  & $W_{down}$ out, layer 28          & 0   & \makecell[r]{key: 12{,}288\\value: 4{,}096} & --- \\
WISE   & $W_{down}$, layer 26 & 0   & 50M & --- \\
\midrule
\multicolumn{5}{@{}l}{\textit{Gemma 3 (12B), 48 LM layers}} \\
\midrule
FT-V   & VL connector      & 4.4M & 0 & --- \\
FT-T   & $W_{down}$, layer 48    & 59M & 0 & --- \\
DPO    & $W_{down}$, layer 48  & 59M & 0 & --- \\
IKE    & ---              & 0   & 0   & SentenceBERT \\
MEND   & FFN, layer 48 & 177M & 0 & Hypernet. (300M) \\
GRACE  & $W_{down}$ out, layer 31          & 0   & \makecell[r]{key: 15{,}360\\value: 3{,}840} & --- \\
WISE   & $W_{down}$, layer 44 & 0   & 59M & --- \\
\bottomrule
\end{tabular}
\caption{Configuration of editing methods per model.
Modified Params count model parameters updated in place; Added Params count new parameters.
IKE retrieves $k=32$ in-context examples per edit, embedded with \texttt{all-MiniLM-L6-v2} SentenceBERT \citep{reimers-2019-sentence-bert}.
MEND's hypernetwork is trained before editing.
GRACE stores a key-value entry \emph{per edit}: when the key (the FFN intermediate state) is matched, the learned value replaces the $W_{down}$ output.
WISE adds a copy of $W_{down}$, updated sparsely per edit (20\% at random).}
\label{tab:edit-config}
\end{table}

All editing evaluations are run on a single NVIDIA H100 80GB GPU, with a total computational budget of approximately 360 GPU hours.

\section{Additional Results}
\subsection{Cross-Modal Evaluation Results} \label{app:cross_eval}

\begin{figure*}
    \includegraphics[width=\textwidth]{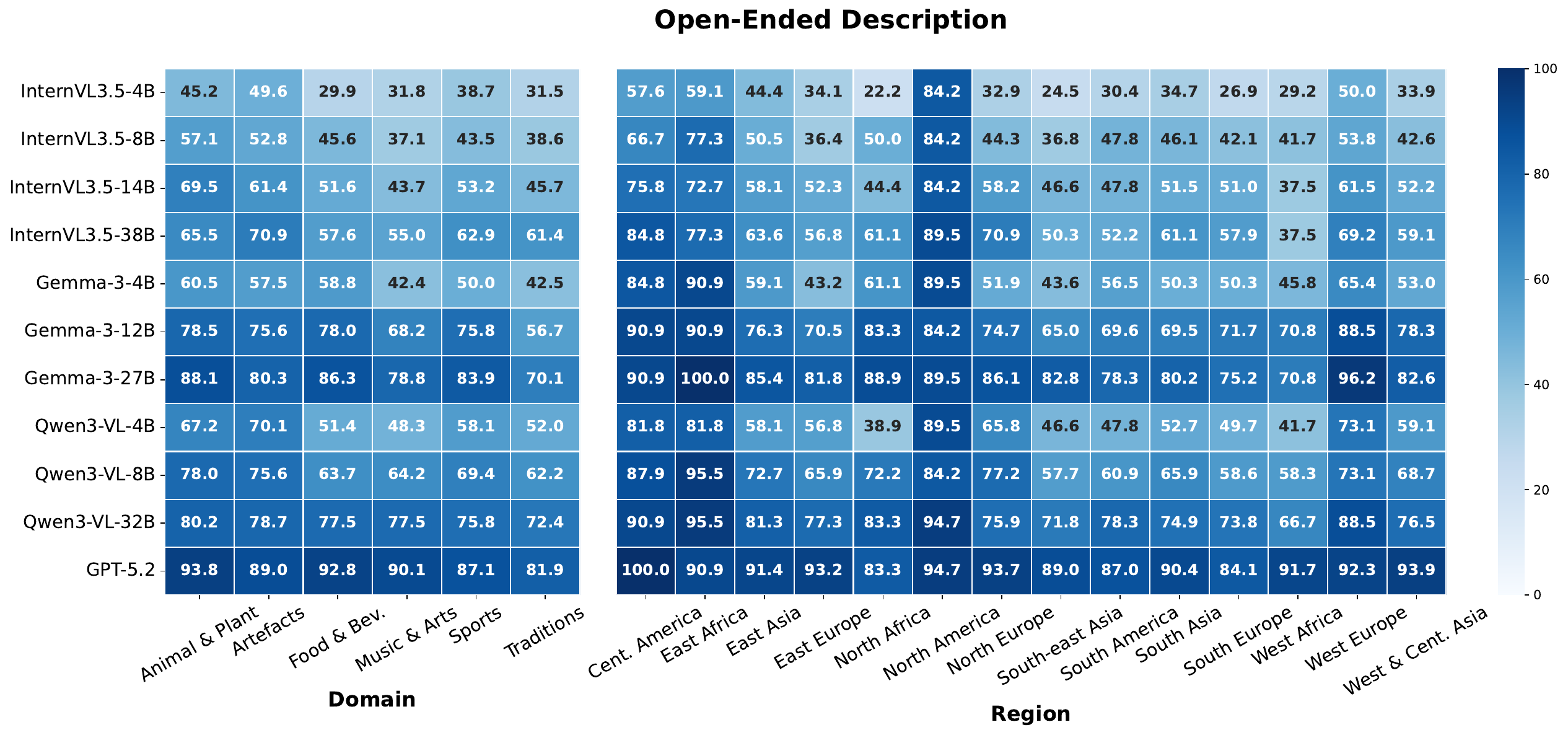}
    \includegraphics[width=\textwidth]{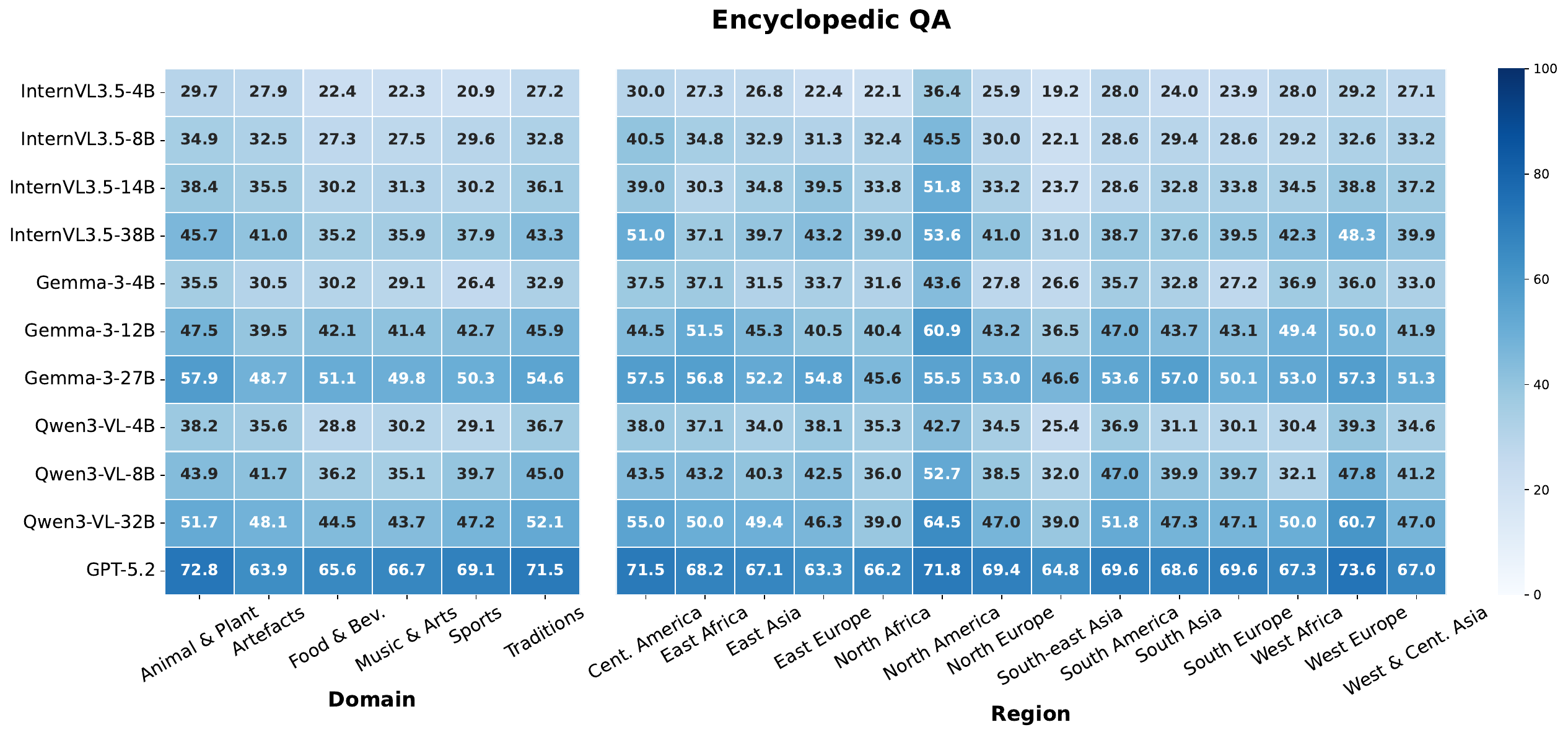}
    \includegraphics[width=\textwidth]{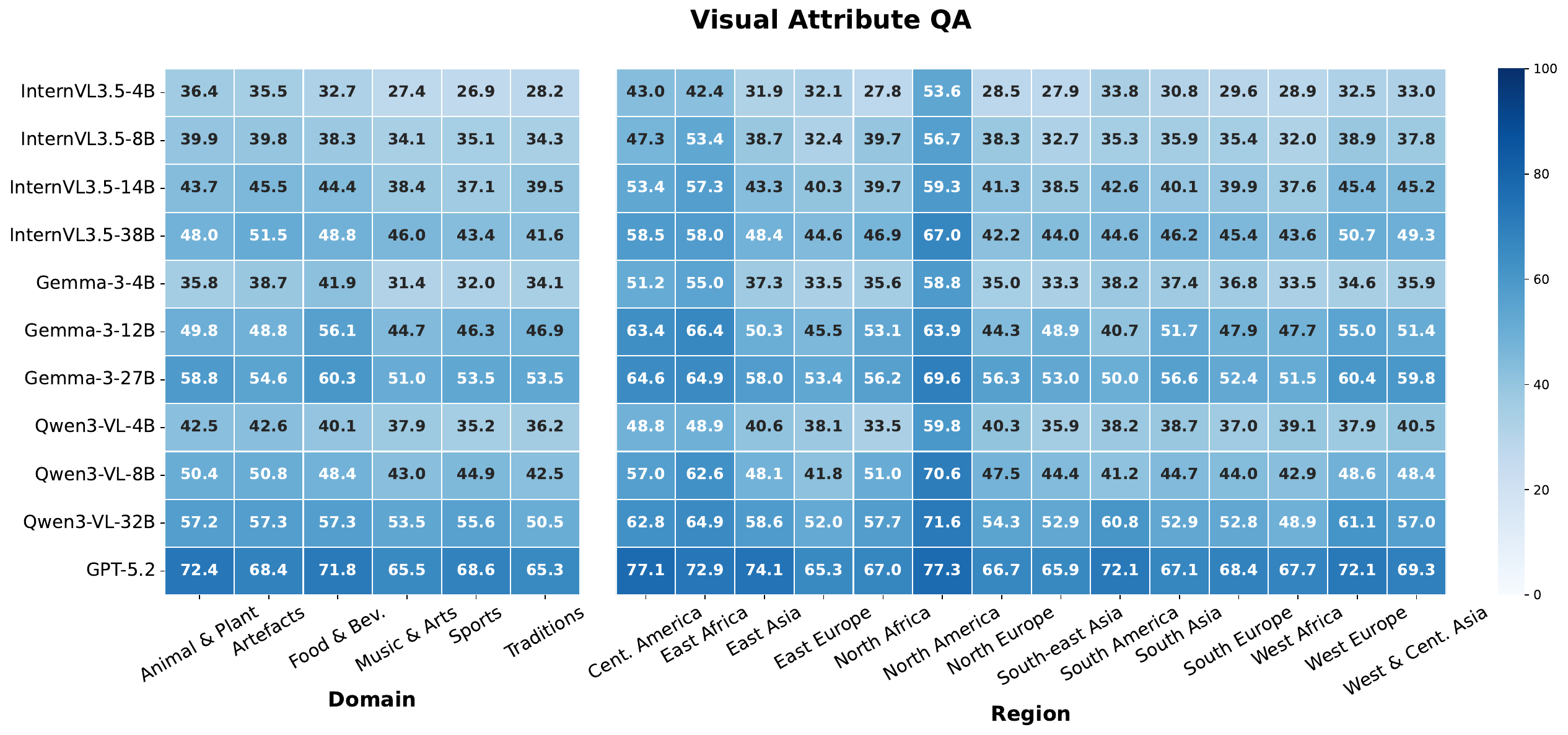}
    \caption{Text-based cross-modal evaluation results per domain and per region.}
    \label{fig:cross_eval_text}
\end{figure*}

\begin{figure*}
    \includegraphics[width=\textwidth]{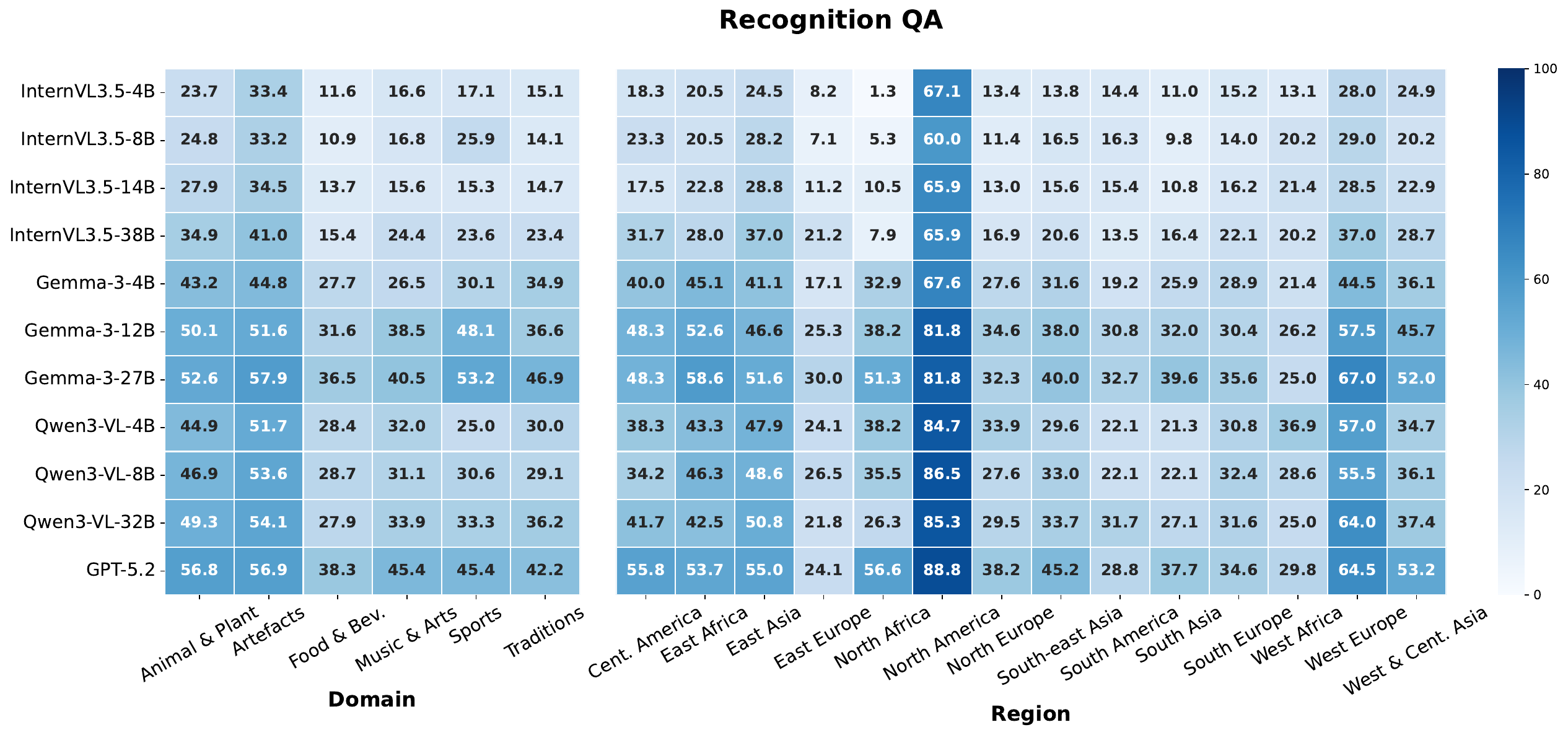}
    \includegraphics[width=\textwidth]{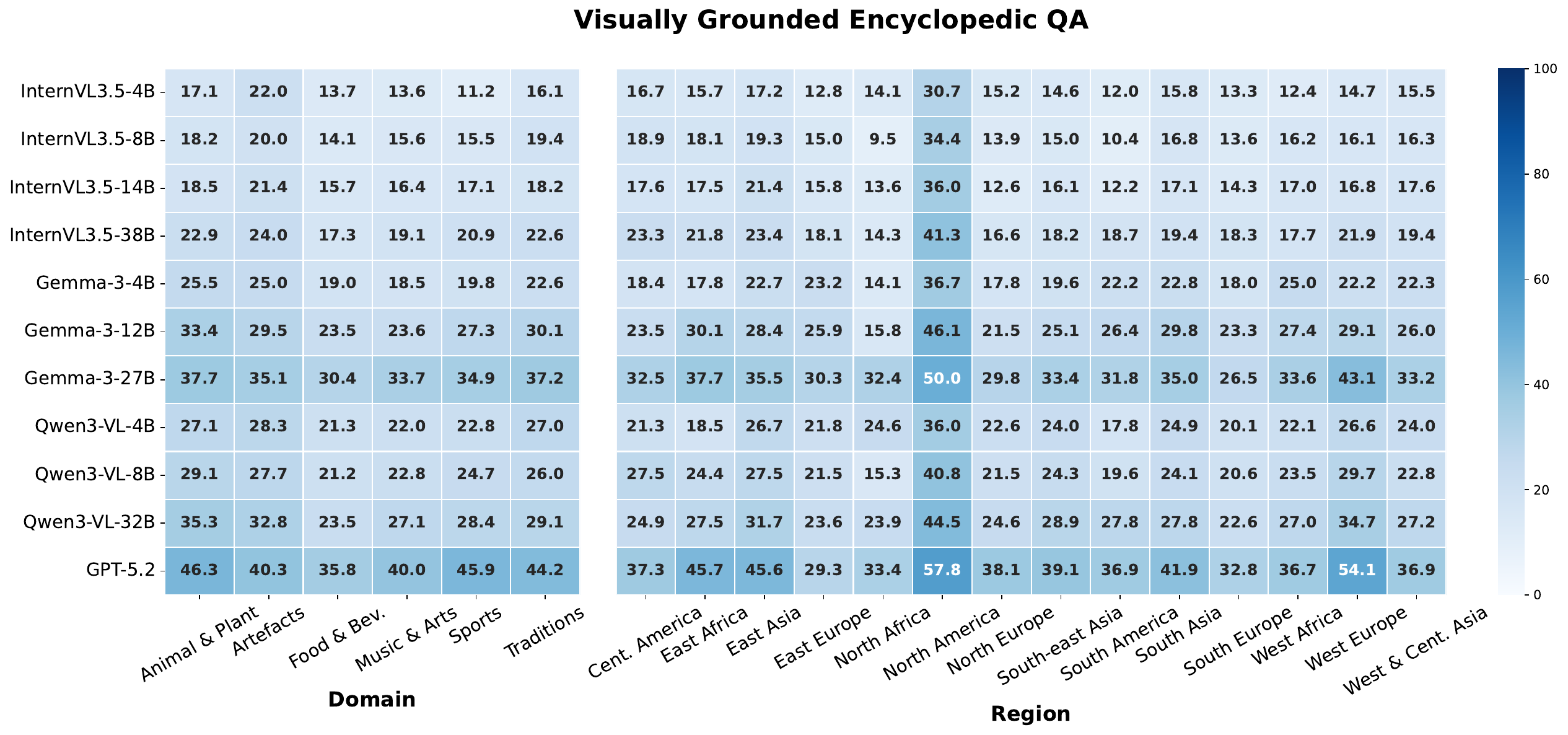}
    \caption{Image-based cross-modal evaluation results per domain and per region.}
    \label{fig:cross_eval_image}
\end{figure*}

\Cref{fig:cross_eval_text} and \Cref{fig:cross_eval_image} present the breakdown of the cross-modal evaluation results per domain and per region.
Across domains, variability is modest for text-based measures but more pronounced for image-based measures, where Foods \& Beverages consistently scores lowest  (mean 24.6\% for Recognition QA), and Animal \& Plant and Artefacts score highest (mean 41.4\% and 46.6\% respectively).
Across regions, we observe consistent patterns.
North America is a persistent outlier, with high performance across models and evaluation dimensions.
Western Europe and East Asia are the next strongest regions, consistent with their higher representation in web-scale pretraining data.
The gap relative to lower-resource regions, such as North Africa, Southeast Asia, and Eastern Europe, is substantial, but the individual weakest regions vary per model family.

\paragraph{Isolating Visual Recognition}
To verify that the drop from Encyclopedic to Visually-Grounded Encyclopedic QA is due to recognition rather than the image input acting as a distractor, we evaluate encyclopedic questions that keep the concept name alongside the image (\Cref{tab:name_image}).
Adding the image to a question that names the concept increases accuracy (+0.4 to +2.1 points), but replacing the name with a referring expression
results to a large drop, supporting that the bottleneck is visual recognition.

\begin{table}[h]
\centering
\small
\setlength{\tabcolsep}{3pt}
\begin{tabular}{lccc}
\toprule
Model & \makecell{Name\\(no image)} & \makecell{Name\\+ image} & \makecell{Ref. expr.\\+ image} \\
\midrule
Qwen3-VL (4B)  & 32.4 & 34.5 & 23.9 \\
Qwen3-VL (8B)  & 39.3 & 39.7 & 24.3 \\
Gemma 3 (12B)  & 43.1 & 45.2 & 26.8 \\
\bottomrule
\end{tabular}
\caption{Encyclopedic QA accuracy when the concept is referenced by name without the image, by name with the image, or by a referring expression with the image (Visually-Grounded Encyclopedic QA).}
\label{tab:name_image}
\end{table}

\subsection{Sequential Edits} \label{app:sequential_edit}

\begin{figure*}[h]
\centering
\begin{subfigure}[t]{0.333\textwidth}       \includegraphics[width=\textwidth]{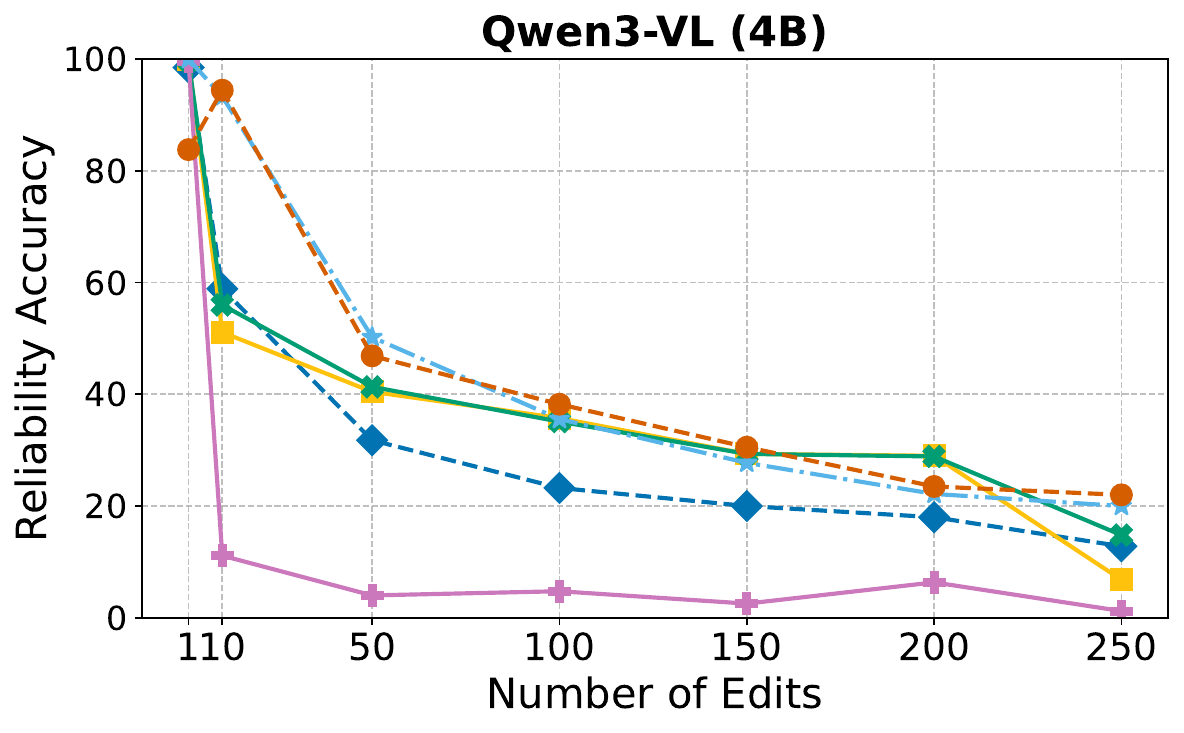}
\end{subfigure}%
\begin{subfigure}[t]{0.333\textwidth}
       \includegraphics[width=\textwidth]{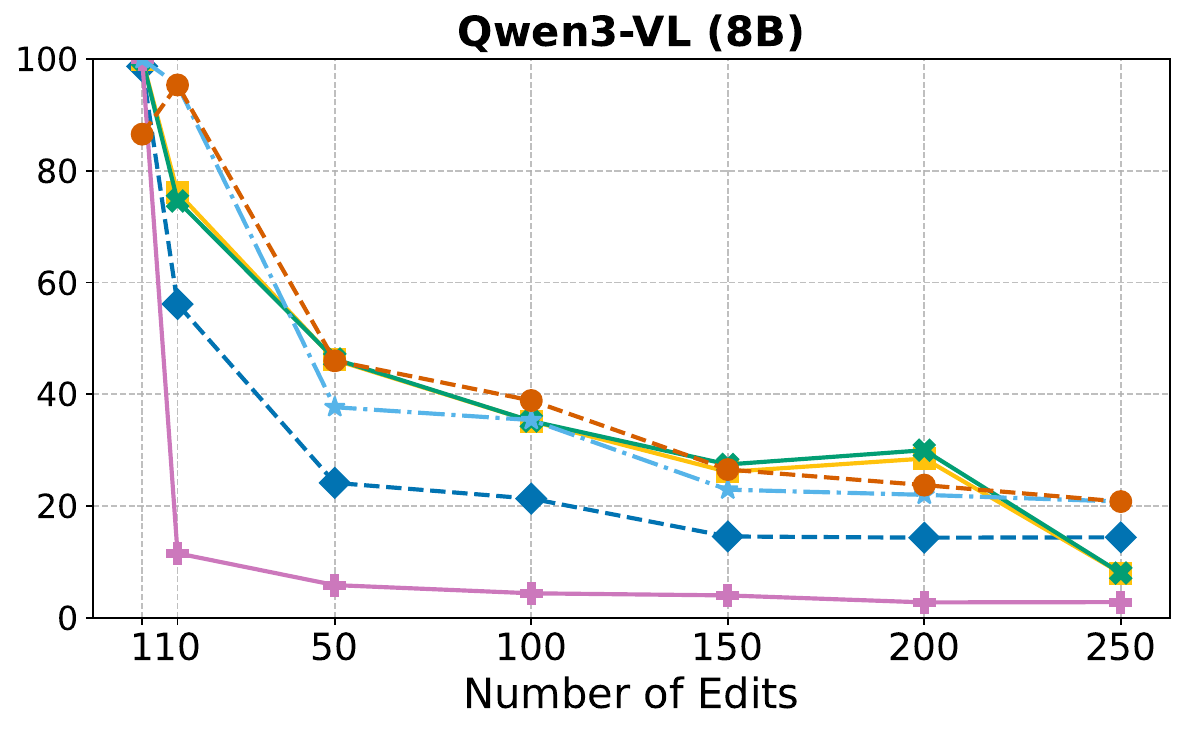}
\end{subfigure}%
\begin{subfigure}[t]{0.333\textwidth}       \includegraphics[width=\textwidth]{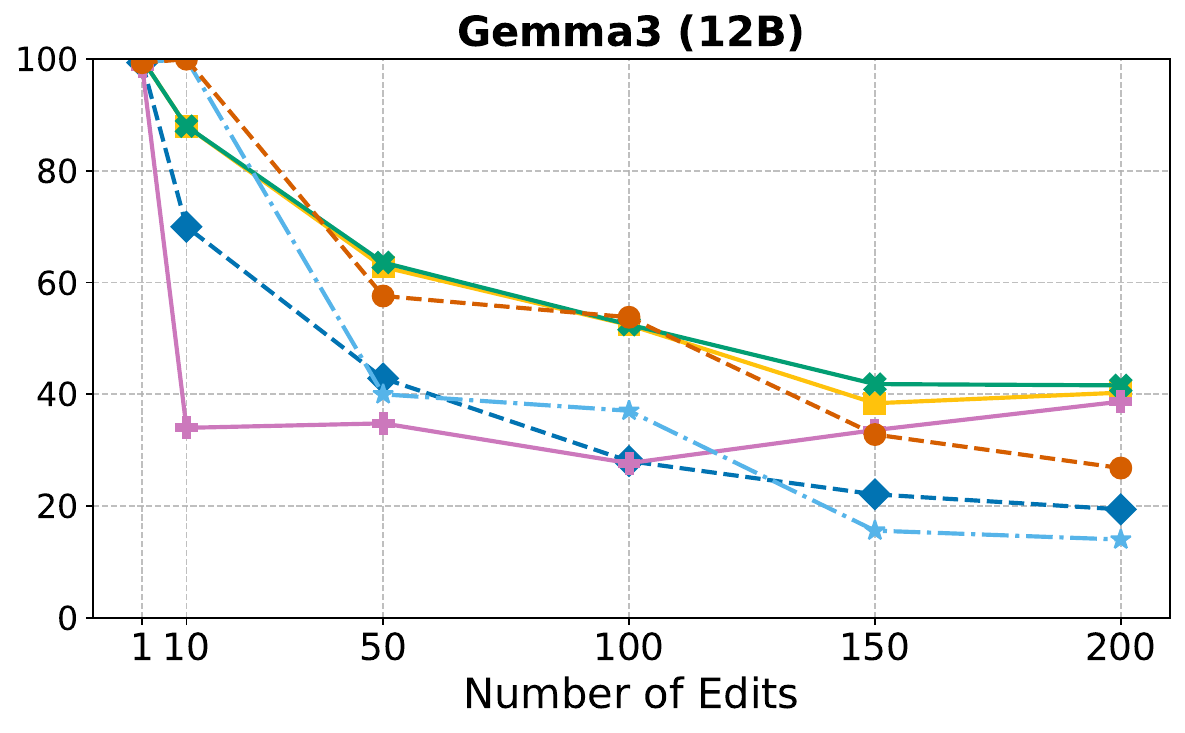}
\end{subfigure}%
\\
\begin{subfigure}[t]{0.333\textwidth}       \includegraphics[width=\textwidth]{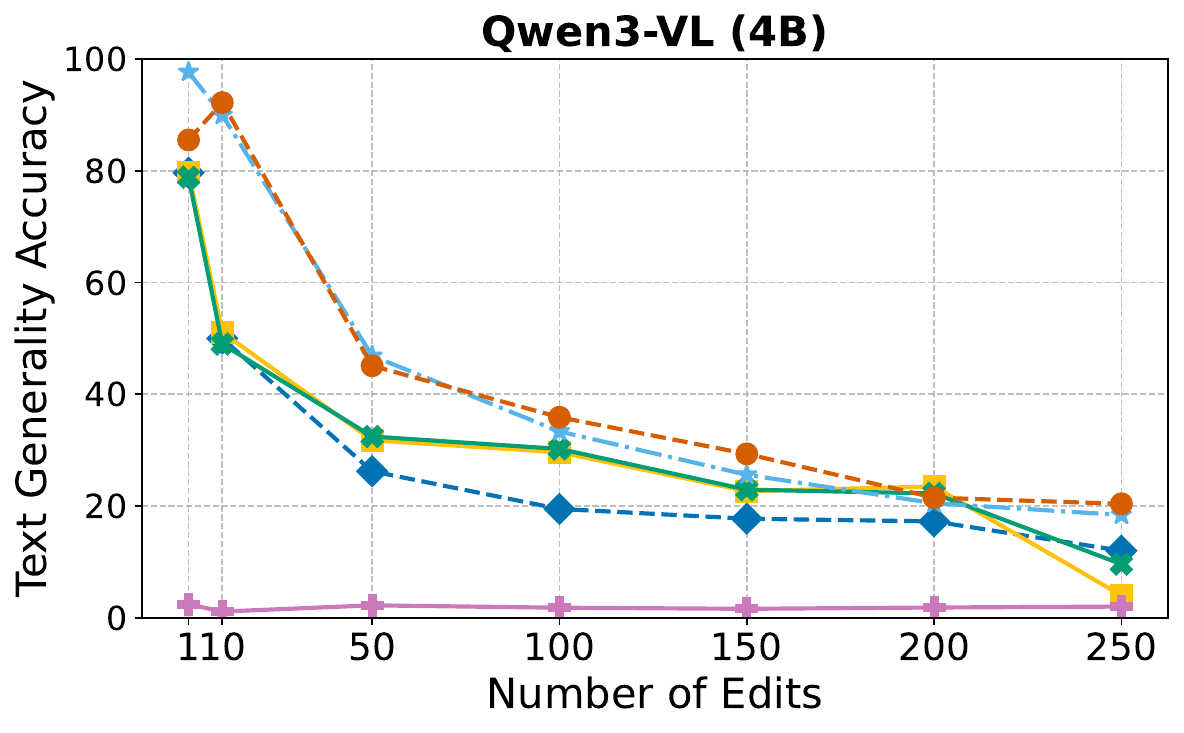}
\end{subfigure}%
\begin{subfigure}[t]{0.333\textwidth}
       \includegraphics[width=\textwidth]{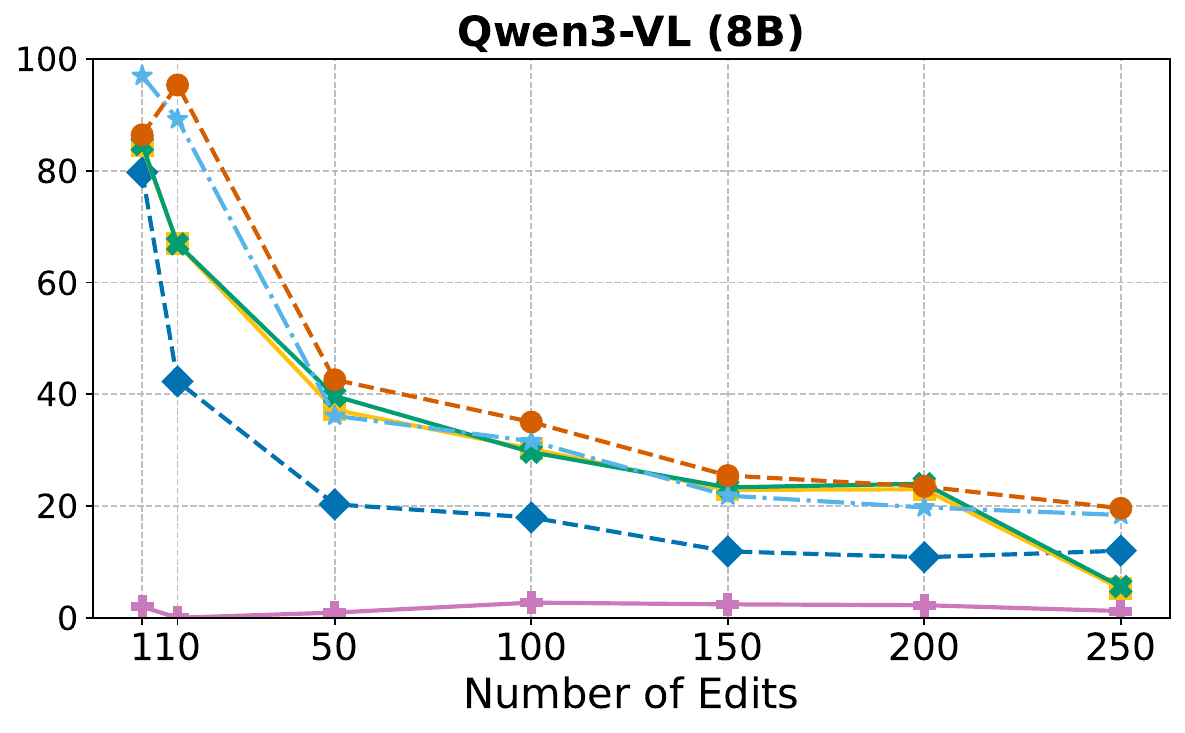}
\end{subfigure}%
\begin{subfigure}[t]{0.333\textwidth}       \includegraphics[width=\textwidth]{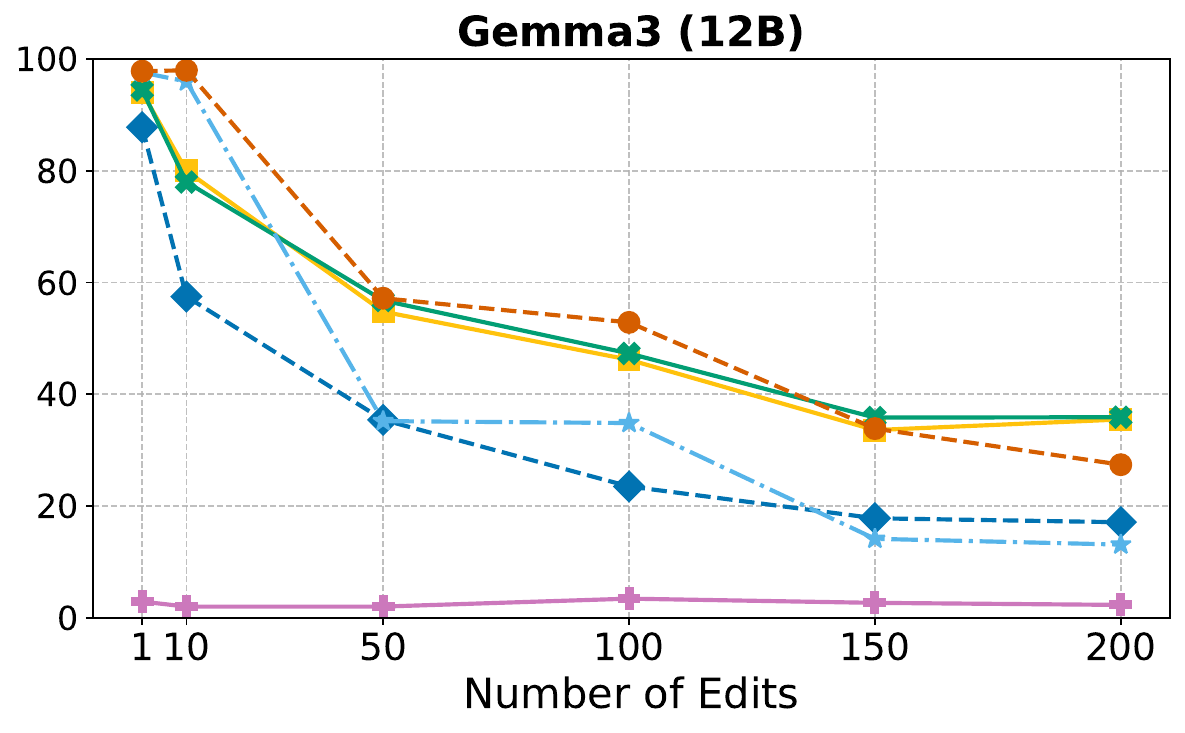}
\end{subfigure}%
\\
\begin{subfigure}[t]{0.333\textwidth}       \includegraphics[width=\textwidth]{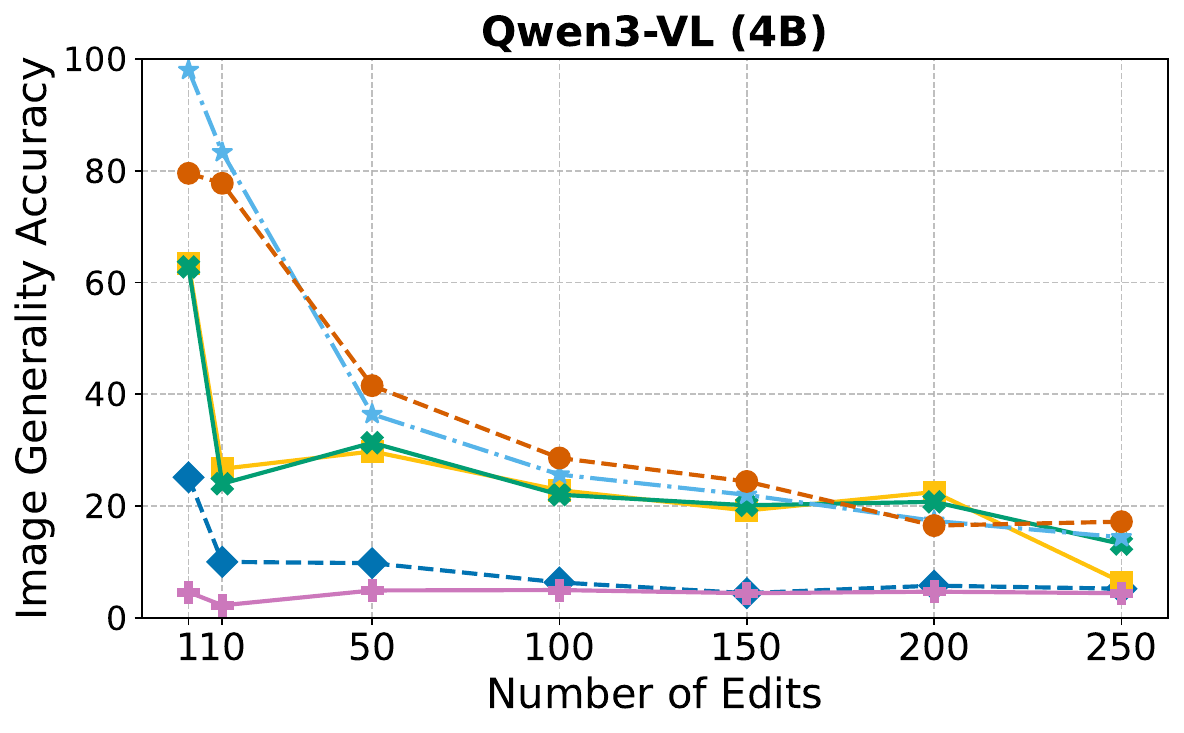}
\end{subfigure}%
\begin{subfigure}[t]{0.333\textwidth}
       \includegraphics[width=\textwidth]{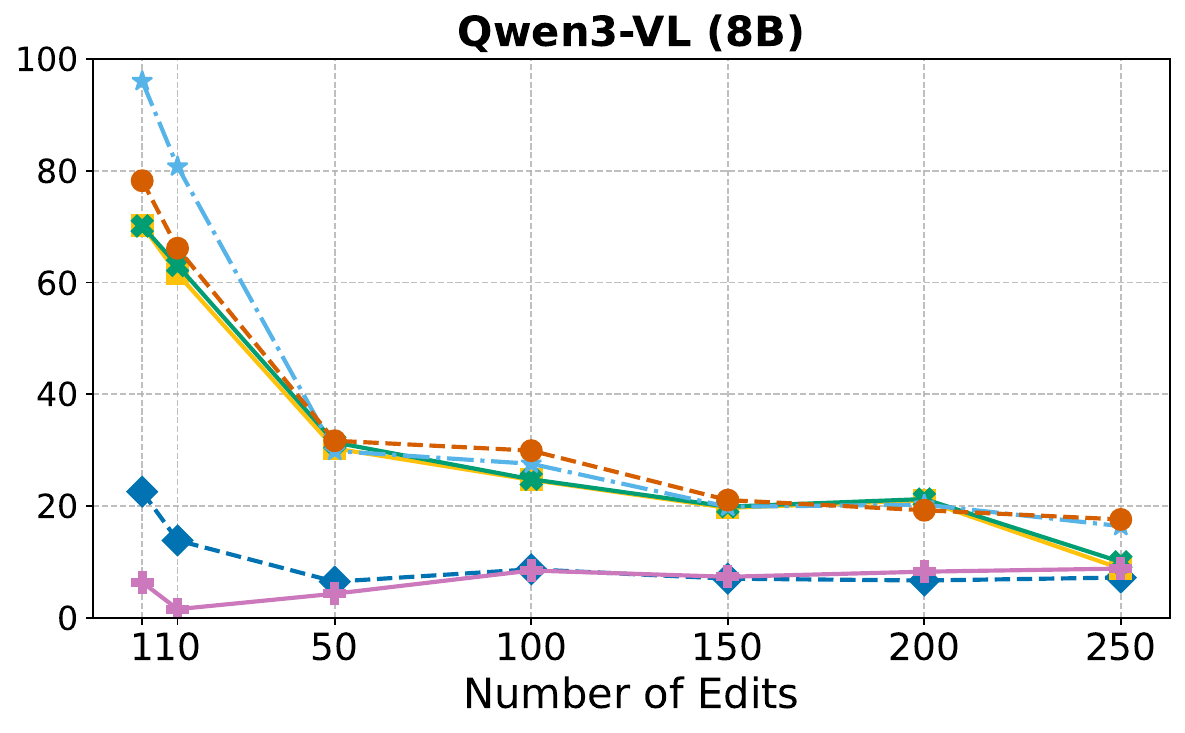}
\end{subfigure}%
\begin{subfigure}[t]{0.333\textwidth}       \includegraphics[width=\textwidth]{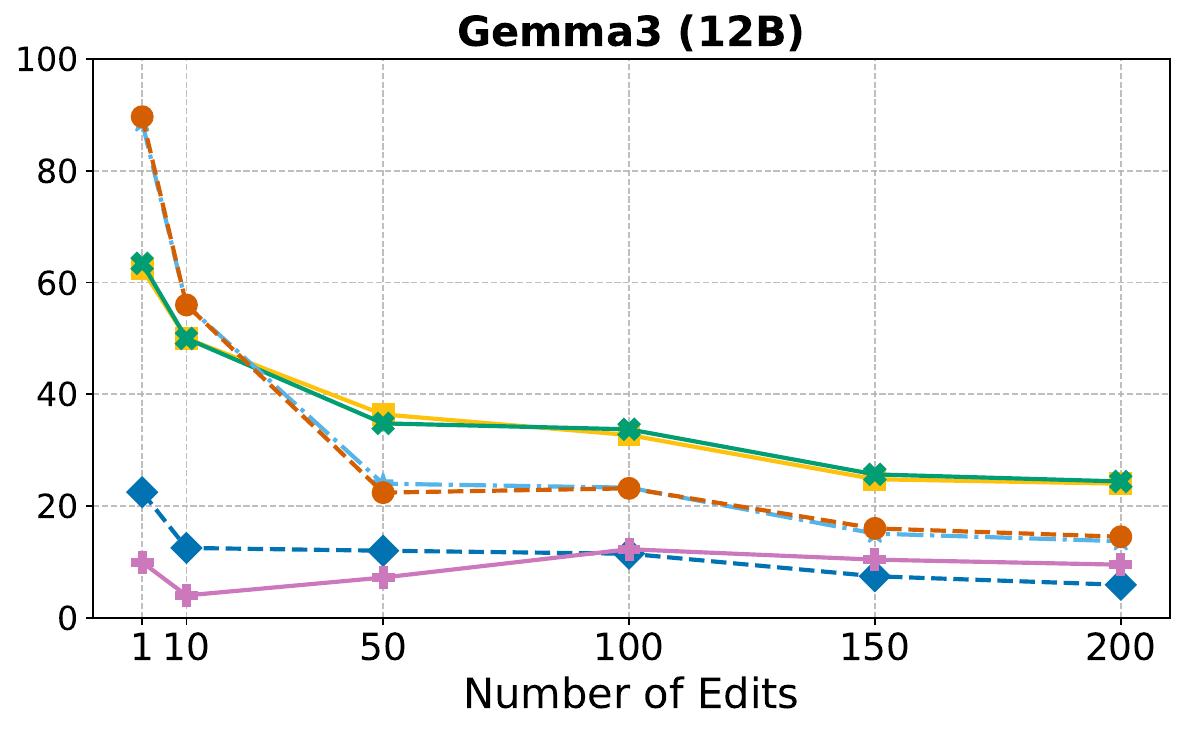}
\end{subfigure}%
\\
\begin{subfigure}[t]{0.333\textwidth}       \includegraphics[width=\textwidth]{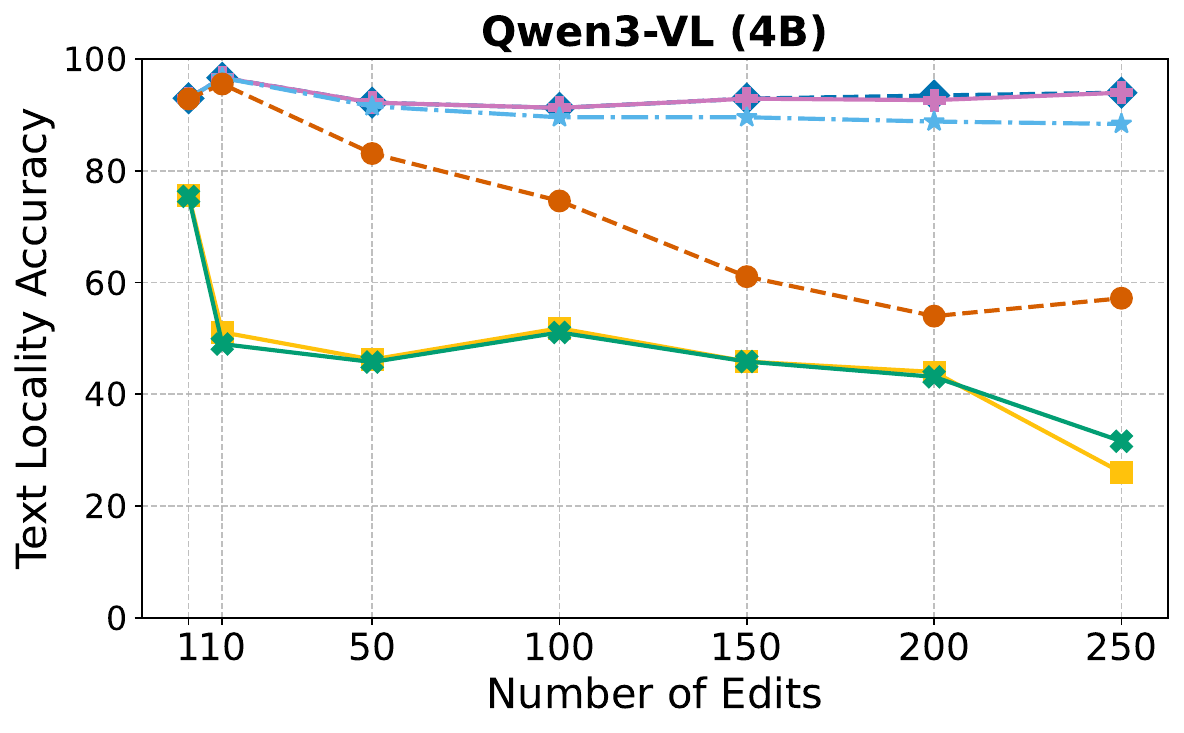}
\end{subfigure}%
\begin{subfigure}[t]{0.333\textwidth}
       \includegraphics[width=\textwidth]{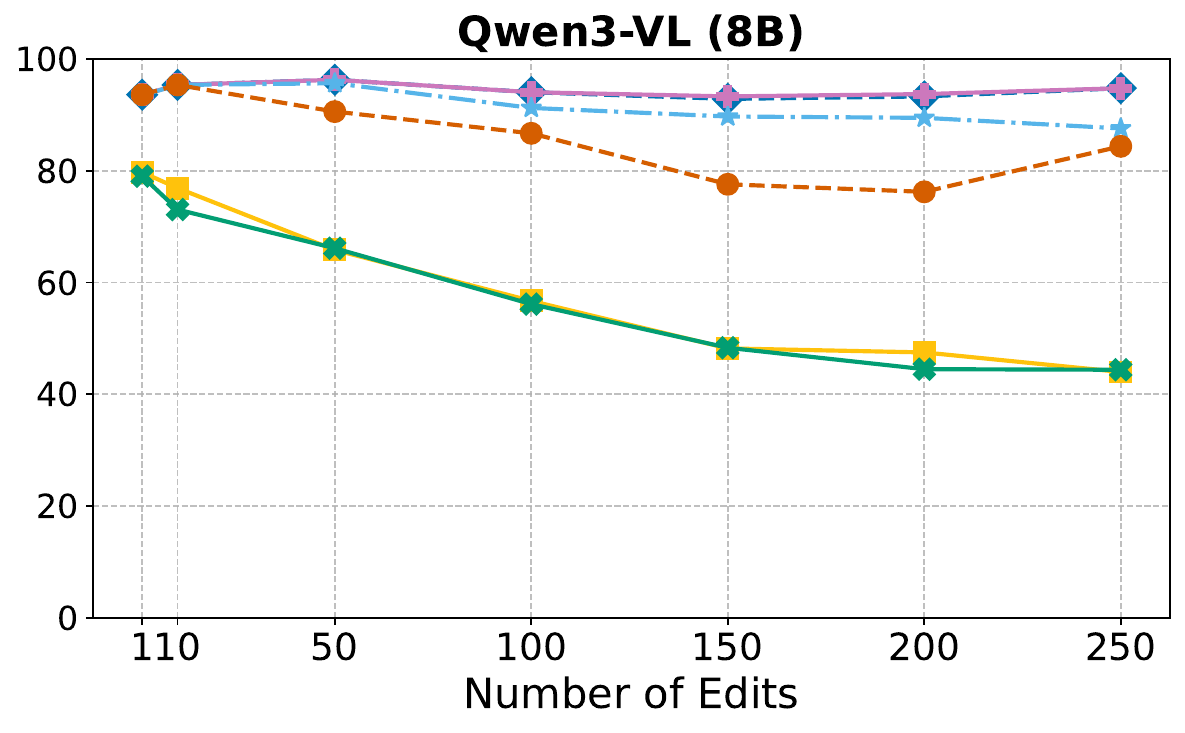}
\end{subfigure}%
\begin{subfigure}[t]{0.333\textwidth}       \includegraphics[width=\textwidth]{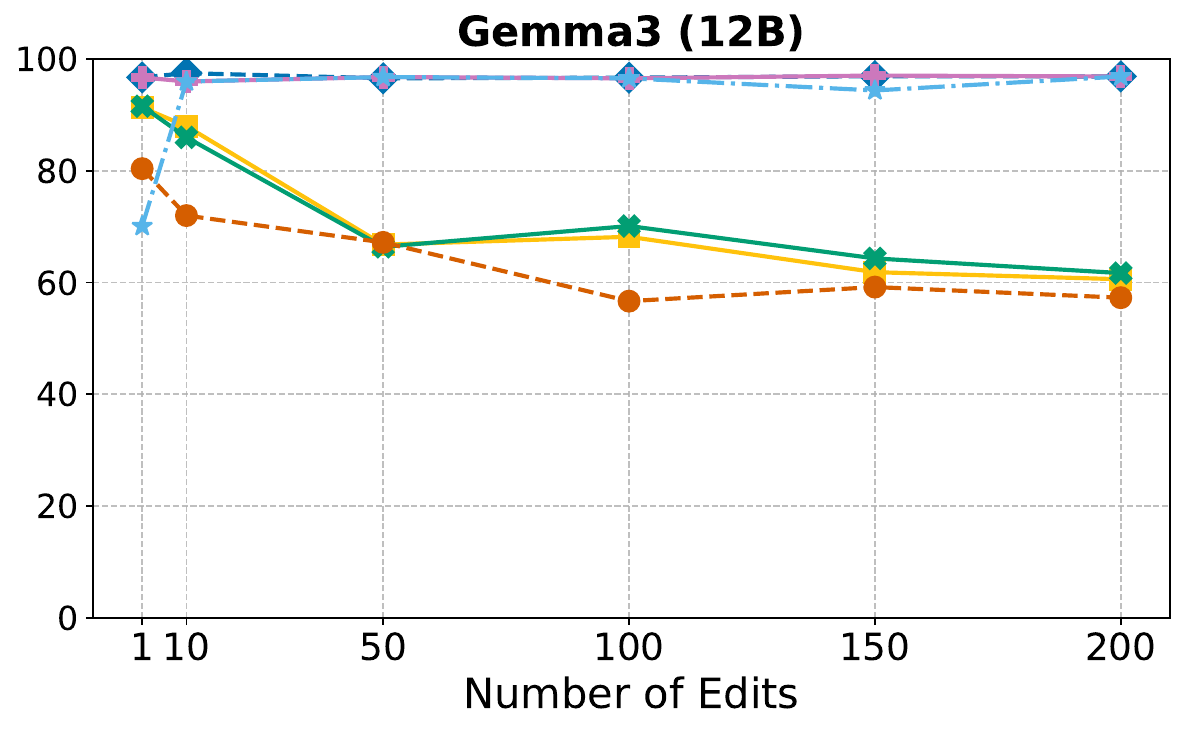}
\end{subfigure}%
\\
\begin{subfigure}[t]{0.333\textwidth}       \includegraphics[width=\textwidth]{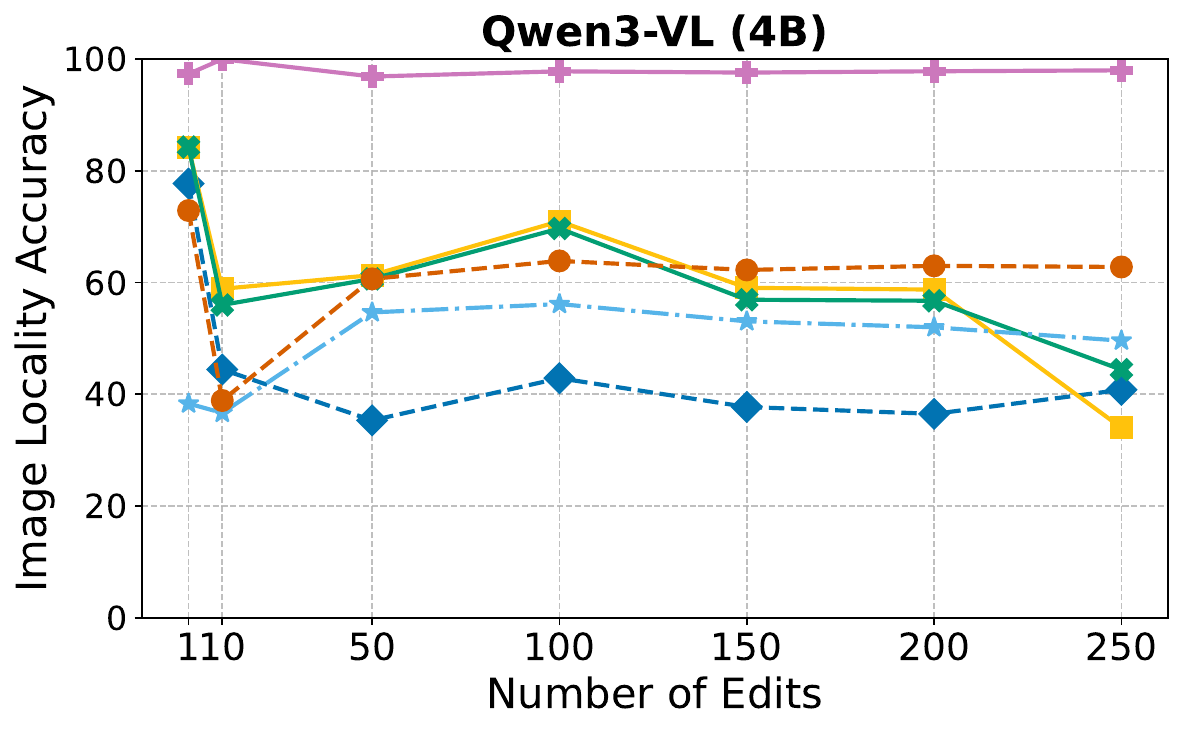}
\end{subfigure}%
\begin{subfigure}[t]{0.333\textwidth}
       \includegraphics[width=\textwidth]{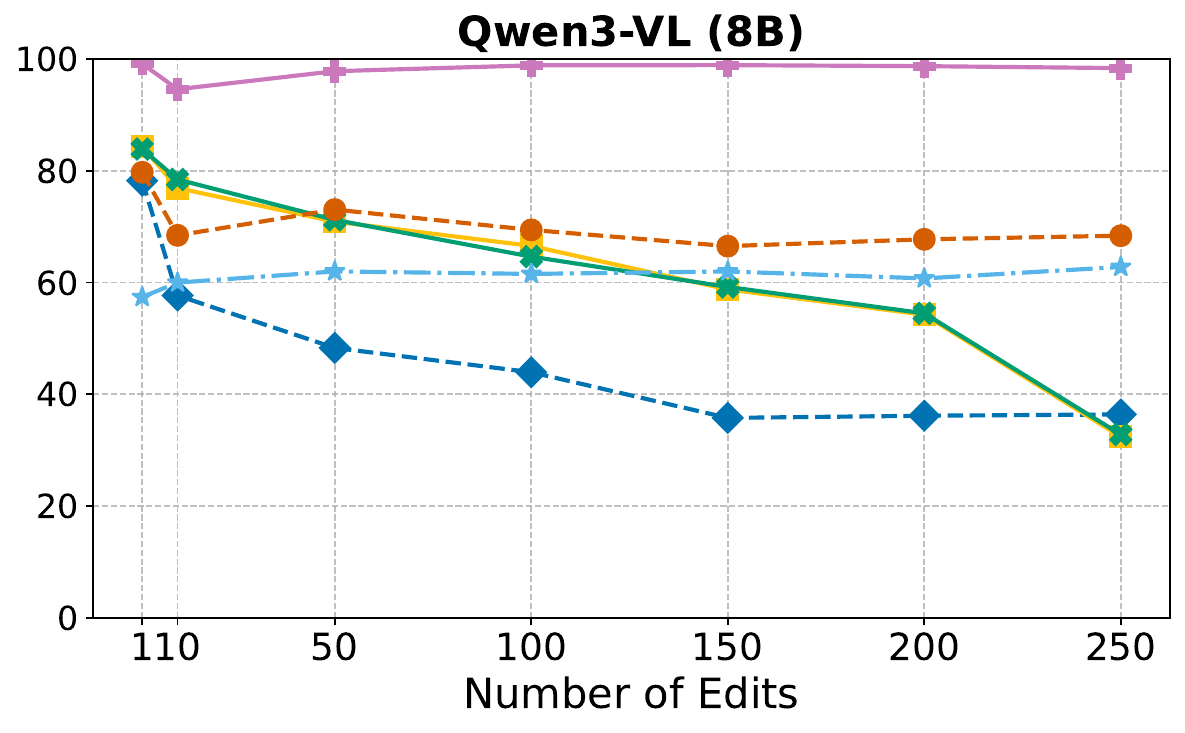}
\end{subfigure}%
\begin{subfigure}[t]{0.333\textwidth}       \includegraphics[width=\textwidth]{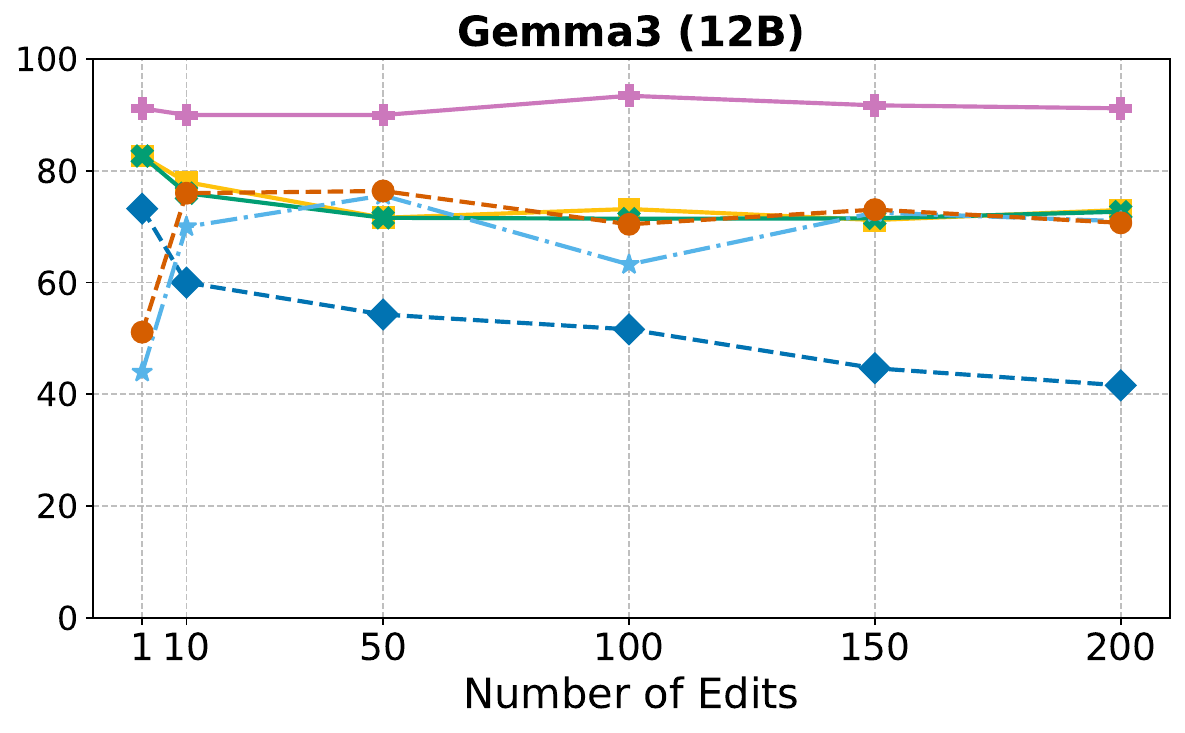}
\end{subfigure}%
\\
\includegraphics[width=0.5\textwidth]{figures/average_accuracy_per_method_Gemma3_12B_legend.pdf}
\caption{Results per editing metric for sequential edits.}
\label{fig:app_sequential_editing_results}
\end{figure*}

As the aggregate view in \Cref{fig:sequential_editing_results} conceals how each method trades off reliability, generality, and locality, we report per-metric results in \Cref{fig:app_sequential_editing_results}.
Reliability and generality are the most impacted across all methods.
Neither finetuning-based nor memory-based approaches sustain either metric beyond the first few edits.
Locality, which requires maintaining unrelated model outputs, is better retained by some methods and varies with the edited component.
Portability is omitted, as it remains uniformly low across methods and edit counts.

FT-T and DPO remain competitive in terms of reliability and generality.
Their text locality declines steadily, potentially due to updates being applied to the final language model layer. FT-V shows the complementary pattern: editing the vision-language connector leaves text locality intact, as text-only inputs bypass the connector, while image locality is the lowest across methods.

Memory-based methods (GRACE and WISE) preserve higher locality across edits.
GRACE preserves near-perfect locality but shows near-zero generality from the first edit onwards. Its flat average performance in \Cref{fig:sequential_editing_results} is, therefore, an artifact of this trade-off rather than evidence of sustained editing success.
The WISE variants perform similarly to direct finetuning on reliability and generality, but each better preserves the locality metric matching the modality used for routing: WISE-T maintains text locality above 85\%, whereas WISE-V retains image locality more consistently.
Note that image locality samples are selected based on their embedding similarity to the edit sample, making this a particularly challenging test of knowledge preservation.

\end{document}